\documentclass[final,5p,times,twocolumn]{elsarticle}

\usepackage{amssymb}
\usepackage{graphicx}
\usepackage{booktabs}
\usepackage{xcolor}
\usepackage{amsmath}
\usepackage{caption}
\usepackage{multirow}
\usepackage{arydshln}
\usepackage{float} 
\usepackage{algorithm} 
\usepackage{bm}
\usepackage{algorithmic} 
\usepackage{amsmath}

\usepackage{dblfloatfix}
\definecolor{darkpink}{RGB}{255,19,166}

\definecolor{matlabgreen}{RGB}{119,172,48}

\usepackage{nomencl}

\definecolor{darkergreen}{rgb}{0.0, 0.5, 0.0}
\makenomenclature

\author[1]{Mehdi Heydari Shahna\corref{cor1}}
\author[1]{Pauli Mustalahti}
\author[1]{Jouni Mattila}

\address[1]{Faculty of Engineering and Natural Sciences, Tampere University, Tampere, 33720, Finland}

\cortext[cor1]{Corresponding author: Mehdi Heydari Shahna\\
E-mail address: mehdi.heydarishahna@tuni.fi }

\begin{document}

\begin{frontmatter}



\title{Robust Vision-based Goal-Reaching Control for Mobile Robots Using a Hierarchical Learning Framework}


\begin{abstract}
Reinforcement learning (RL) has strong potential in robotics, but exploration-based training complicates safe deployment on large-scale robots. For such applications, this paper proposes a novel hierarchical goal-reaching framework that integrates stereo visual pose estimation, constrained RL-based motion planning, actuator-level robust adaptive control (RAC), and supervisory safe-return logic. Stereo visual localization is used as the real-time pose-estimation interface with loop closing, map fusion, and relocalization. The RL planner generates smooth, feasible goal-reaching references using a problem-specific reward structure and motion constraints that promote goal progress, reduce oscillations, preserve vision-consistent smoothness, and respect the mechanical limits of a heavy skid-steered robot. At the actuation layer, a scaled conjugate-gradient (SCG)-trained deep neural network (DNN) approximates a quasi-static actuator feedforward map from wheel-speed data to nominal control input. This feedforward map is combined with a logarithmic-barrier-based RAC to compensate for residual modeling errors, slip-induced disturbances, and bounded mismatch between the nominal map and real actuator response. For the actuator-level wheel-tracking subsystem, uniformly ultimately bounded tracking with exponential convergence to a disturbance-dependent residual set is established under bounded uncertainty. A logarithmic safety supervisor monitors execution, detects unsafe operating conditions, including faults and localization inconsistencies, and switches the robot to safe-return mode. Experiments on a 6000 kg robot over asphalt and loose-soil terrain demonstrate approximately 3–4 cm final-position root mean square error (RMSE), accurate tracking of RL-generated commands, improved actuator-level performance over two RAC baselines, and successful autonomous recovery after fault injection.
\end{abstract}


\end{frontmatter}

\section*{Nomenclature}
\vspace{-0.8cm}
\printnomenclature

\nomenclature[A01]{\textit{Abbreviations}}{}
\nomenclature[A02]{\textbf{MDP}}{Markov decision process}
\nomenclature[A03]{\textbf{RL}}{Reinforcement learning}
\nomenclature[A04]{\textbf{SLAM}}{Simultaneous localization and mapping}
\nomenclature[A05]{\textbf{ORB-SLAM3}}{Oriented FAST and Rotated BRIEF SLAM, version 3}
\nomenclature[A06]{\textbf{DNN}}{Deep neural network}
\nomenclature[A07]{\textbf{SCG}}{Scaled conjugate gradient}
\nomenclature[A08]{\textbf{RAC}}{Robust adaptive controller}
\nomenclature[A09]{\textbf{TD}}{Temporal difference}
\nomenclature[A10]{\textbf{RMSE}}{Root mean square error}
\nomenclature[A11]{\textbf{GNSS}}{Global navigation satellite system}
\nomenclature[A12]{\textbf{PMSM}}{Permanent magnet synchronous motor}
\nomenclature[A13]{\textbf{CBF}}{Control barrier function}
\nomenclature[A14]{\textbf{APE}}{Absolute pose error}
\nomenclature[A15]{\textbf{RPE}}{Relative pose error}
\nomenclature[A16]{\textbf{INS-RTK}}{Inertial navigation system with real-time kinematic GNSS}
\nomenclature[A17]{\textbf{RGB}}{Red-green-blue}
\nomenclature[A18]{\textbf{ORB}}{Oriented FAST and Rotated BRIEF}
\nomenclature[A19]{\textbf{FAST}}{Features from Accelerated Segment Test}
\nomenclature[A20]{\textbf{DBoW2}}{Bag-of-Words library, version 2}
\nomenclature[A21]{\textbf{EPnP}}{Efficient Perspective-n-Point}
\nomenclature[A22]{\textbf{MSE}}{Mean squared error}
\nomenclature[A23]{\textbf{MAE}}{Mean absolute error}
\nomenclature[A24]{\textbf{LSTM}}{Long short-term memory}
\nomenclature[A25]{\textbf{GRU}}{Gated recurrent unit}
\nomenclature[A26]{\textbf{ReLU}}{Rectified linear unit}

\nomenclature[B01]{\textit{Sets and operators}}{}
\nomenclature[B02]{$\mathbb{R}$}{Set of real numbers}
\nomenclature[B03]{$\mathbb{N}$}{Set of natural numbers}
\nomenclature[B04]{$SE(3)$}{Special Euclidean group of rigid-body poses}
\nomenclature[B05]{$SO(3)$}{Special orthogonal group of rotation matrices}
\nomenclature[B06]{$\mathbb{E}_{\pi}[\cdot]$}{Expectation under policy $\pi$}
\nomenclature[B07]{$\mathrm{sat}(\cdot)$}{Saturation operator}
\nomenclature[B08]{$\mathrm{wrap}_{\pi}(\cdot)$}{Angle-wrapping operator to $[-\pi,\pi)$}
\nomenclature[B09]{$\mathrm{atan2}(\cdot,\cdot)$}{Four-quadrant inverse tangent}
\nomenclature[B10]{$\mathrm{sign}(\cdot)$}{Sign function}
\nomenclature[B11]{$\odot$}{Elementwise product}

\nomenclature[C01]{\textit{SLAM and RL motion planning}}{}
\nomenclature[C02]{$\mathcal{M}$}{Finite MDP, $\mathcal{M}=(\mathcal{S},\mathcal{A},P,\mathcal{R},\gamma)$}
\nomenclature[C03]{$\mathcal{S}$}{State space}
\nomenclature[C04]{$\mathcal{A}$}{Action space}
\nomenclature[C05]{$P(s'|s,a)$}{State-transition probability kernel}
\nomenclature[C06]{$\mathcal{R}(s_t,a_t,s_{t+1})$}{Reward function}
\nomenclature[C07]{$\gamma$}{Discount factor}
\nomenclature[C08]{$\pi(a|s)$}{Policy}
\nomenclature[C09]{$T$}{Camera pose, $T=[R_c,x_{\mathrm{msr}}]\in SE(3)$}
\nomenclature[C10]{$R_c$}{Camera rotation matrix in $SO(3)$}
\nomenclature[C11]{$x_{\mathrm{msr}}$}{Measured pose vector, $[x,y,\theta]^\top$}
\nomenclature[C12]{$x_t,\;y_t$}{Robot planar position at time step $t$}
\nomenclature[C13]{$\theta_t$}{Robot heading angle at time step $t$}
\nomenclature[C14]{$v_t$}{Robot linear velocity at time step $t$}
\nomenclature[C15]{$\omega_t$}{Robot angular velocity at time step $t$}
\nomenclature[C16]{$x_g,\;y_g$}{Goal coordinates}
\nomenclature[C17]{$d_t$}{Distance to the goal}
\nomenclature[C18]{$e_t$}{Heading error}
\nomenclature[C19]{$d^{(i)},\,e^{(i)},\,v^{(i)},\,\omega^{(i)}$}{Bin boundaries or representative values}
\nomenclature[C20]{$N_d,\;N_{\theta},\;N_v,\;N_{\omega}$}{Numbers of distance, heading, linear-velocity, and angular-velocity bins}
\nomenclature[C21]{$d_{\max}$}{Maximum distance used for discretization}
\nomenclature[C22]{$s_t=(i_d,i_e,i_v,i_{\omega})$}{Discrete RL state}
\nomenclature[C23]{$a_t=(a_{v,t},a_{\omega,t})$}{Acceleration action}
\nomenclature[C24]{$a_{v,t}$}{Linear-acceleration action}
\nomenclature[C25]{$a_{\omega,t}$}{Angular-acceleration action}
\nomenclature[C26]{$a_v^{(i)},\,a_{\omega}^{(i)}$}{Discrete acceleration levels}
\nomenclature[C27]{$v_{\min},\,v_{\max}$}{Linear-speed limits}
\nomenclature[C28]{$\omega_{\min},\,\omega_{\max}$}{Angular-speed limits}
\nomenclature[C29]{$a_{v,\min},\,a_{v,\max}$}{Linear-acceleration limits}
\nomenclature[C30]{$a_{\omega,\min},\,a_{\omega,\max}$}{Angular-acceleration limits}
\nomenclature[C31]{$\Delta t$}{Sampling time step}
\nomenclature[C32]{$T_{\mathrm{ep}}$}{Episode terminal time}
\nomenclature[C33]{$J^{\pi}$}{Expected discounted return}
\nomenclature[C34]{$Q(s,a)$}{Tabular action-value function}
\nomenclature[C35]{$\alpha$}{Learning rate}
\nomenclature[C36]{$\delta_t$}{Temporal-difference error}
\nomenclature[C37]{$\varepsilon_t$}{Exploration probability in the $\varepsilon$-greedy policy}
\nomenclature[C38]{$\varepsilon_0,\;\varepsilon_{\mathrm{final}}$}{Initial and final exploration probabilities}
\nomenclature[C39]{$\mathcal{P}$}{State-transition probability kernel}
\nomenclature[C40]{\ensuremath{N_{a_v}, N_{a_\omega}}}{Numbers of linear- and angular-acceleration levels}
\nomenclature[C40]{$N_{a_v}, N_{a_\omega}$}{Numbers of linear- and angular-acceleration levels}
\nomenclature[C41]{$x_{\mathrm{msr}}$}{Measured planar position vector, $x_{\mathrm{msr}}=[x,y]^\top$}
\nomenclature[C42]{$n_j$}{Number of neurons in layer $j$}
\nomenclature[C43]{$\beta_i^\star$}{Selected trained DNN parameter vector for wheel $i$}
\nomenclature[C44]{$k_{\omega,\mathrm{stop}}$}{Predictive stopping-angle penalty weight}
\nomenclature[C45]{$k_{\omega,\mathrm{sign}}$}{Wrong-sign angular penalty weight}
\nomenclature[D01]{\textit{Reward terms and RL parameters}}{}
\nomenclature[D02]{$r_t$}{Immediate reward}
\nomenclature[D03]{$R_{\mathrm{task}}$}{Task-reward component}
\nomenclature[D04]{$R_{\mathrm{shape}}$}{Reward-shaping component}
\nomenclature[D05]{$r_{\mathrm{step}}$}{Step-cost term}
\nomenclature[D06]{$r_d$}{Distance-progress term}
\nomenclature[D07]{$r_{\mathrm{timeout}}$}{Timeout penalty term}
\nomenclature[D08]{$r_{\theta}$}{Heading-progress term}
\nomenclature[D09]{$r_{\omega}$}{Angular-velocity penalty term}
\nomenclature[D10]{$r_{\parallel v}$}{Forward-motion shaping term}
\nomenclature[D11]{$r_{\perp v}$}{Lateral-motion penalty term}
\nomenclature[D12]{$r_a$}{Acceleration penalty term}
\nomenclature[D13]{$r_{\mathrm{hyst}}$}{Hysteresis penalty term}
\nomenclature[D14]{$r_{\mathrm{goal}}$}{Goal bonus term}
\nomenclature[D15]{$r_{\mathrm{flip}}$}{Angular sign-flip penalty}
\nomenclature[D16]{$r_{\mathrm{inc}}$}{Heading-increase penalty}
\nomenclature[D17]{$r_{\mathrm{stall}}$}{Heading-stall penalty}
\nomenclature[D18]{$r_{\mathrm{stop}}$}{Predictive stopping penalty}
\nomenclature[D19]{$r_{\mathrm{sign}}$}{Wrong-sign angular penalty}
\nomenclature[D20]{$k_{\mathrm{step}}$}{Step-cost weight}
\nomenclature[D21]{$k_d$}{Distance-progress weight}
\nomenclature[D22]{$k_{\mathrm{timeout}}$}{Timeout weight}
\nomenclature[D23]{$k_{\theta}$}{Heading-progress weight}
\nomenclature[D24]{$k_{\omega}$}{Angular-velocity penalty weight}
\nomenclature[D25]{$k_v$}{Forward-motion shaping weight}
\nomenclature[D26]{$k_{\mathrm{lat}}$}{Lateral-motion penalty weight}
\nomenclature[D27]{$k_{a_v},\,k_{a_{\omega}}$}{Acceleration penalty weights}
\nomenclature[D28]{$k_{\mathrm{ws}}$}{Hysteresis penalty weight}
\nomenclature[D29]{$k_{\omega\mathrm{flip}}$}{Angular sign-flip penalty weight}
\nomenclature[D30]{$k_{\mathrm{head,inc}}$}{Heading-increase penalty weight}
\nomenclature[D31]{$k_{\mathrm{head,stall}}$}{Heading-stall penalty weight}
\nomenclature[D32]{$k_{\omega\mathrm{stop}}$}{Stopping-angle penalty weight}
\nomenclature[D33]{$k_{\omega\mathrm{sign}}$}{Wrong-sign angular penalty weight}
\nomenclature[D34]{$d_T$}{Distance to the goal at timeout}
\nomenclature[D35]{$e_{\mathrm{db}}$}{Heading deadband}
\nomenclature[D36]{$\omega_{\mathrm{db}}$}{Angular-velocity deadband}
\nomenclature[D37]{$d_{\mathrm{goalTol}}$}{Goal tolerance}
\nomenclature[D38]{$e_{\mathrm{lock}}$}{Heading lock threshold}
\nomenclature[D39]{$d_{\mathrm{lock}}$}{Distance lock threshold}
\nomenclature[D40]{$\Delta e$}{Change in heading-error magnitude, $\Delta e=|e_{t+1}|-|e_t|$}
\nomenclature[D41]{$\theta_{\mathrm{stop}}$}{Predicted stopping angle}
\nomenclature[D42]{$a_{\omega}^{B}$}{Angular braking bound}
\nomenclature[D43]{$e_{\mathrm{pad}}$}{Stopping-angle safety margin}
\nomenclature[D44]{$s_0$}{Initial sign of the heading error}
\nomenclature[D45]{$\Phi(s_t)$}{Potential function}
\nomenclature[D48]{$\Delta e$}{Heading-error magnitude change, $\Delta e=|e_{t+1}|-|e_t|$}
\nomenclature[D49]{$\Delta e_{\mathrm{head}}$}{Heading-stall threshold}
\nomenclature[D50]{$\mathrm{excess}$}{Predictive stopping-angle violation}
\nomenclature[D51]{$\mathrm{wrong}$}{Wrong-sign angular-velocity residual}

\nomenclature[E01]{\textit{DNN actuator feedforward map}}{}
\nomenclature[E02]{$r$}{Wheel radius}
\nomenclature[E03]{$\omega_i$}{Angular speed of wheel $i$}
\nomenclature[E04]{$v_i$}{Tangential wheel speed, $v_i=r\omega_i$}
\nomenclature[E05]{$u_i$}{Applied control input of wheel $i$}
\nomenclature[E06]{$\hat{u}_i$}{DNN-predicted control input}
\nomenclature[E07]{$f_{\beta}(\cdot)$}{DNN feedforward map from wheel speed to nominal control input}
\nomenclature[E08]{$\beta$}{DNN parameter vector}
\nomenclature[E09]{$\beta_i$ }{DNN parameter vector for wheel $i$}
\nomenclature[E10]{$L$}{Number of hidden layers}
\nomenclature[E11]{$a^{(j)}$}{Activation vector at layer $j$}
\nomenclature[E12]{$z^{(j)}$}{Pre-activation vector at layer $j$}
\nomenclature[E13]{$W^{(j)}$}{Weight matrix of layer $j$}
\nomenclature[E14]{$b^{(j)}$}{Bias vector of layer $j$}
\nomenclature[E15]{$\phi(\cdot)$}{Activation function}
\nomenclature[E16]{$\phi'(\cdot)$}{Derivative of the activation function}
\nomenclature[E17]{$N_{\mathrm{train}}$}{Number of training samples}
\nomenclature[E18]{$J(\beta_i)$ }{Training loss function for wheel $i$}
\nomenclature[E25]{$\beta^{\star}$}{Selected DNN parameter vector for wheel \(i\) after training}
\nomenclature[E26]{$u_{\mathrm{FF},i}$}{DNN feedforward control input for wheel $i$}
\nomenclature[E27]{$A_i$}{Unknown positive actuator coefficient}

\nomenclature[F01]{\textit{RAC and safety}}{}
\nomenclature[F02]{$v_{d,i}$}{Desired speed of wheel $i$}
\nomenclature[F04]{$v_d$}{Desired wheel-speed vector}
\nomenclature[F05]{$e_i$}{Wheel-speed tracking error, $e_i=v_i-v_{d,i}$}
\nomenclature[F06]{$d_i(v_{d,i},e_i,t)$}{Lumped uncertainty or disturbance term}
\nomenclature[F07]{$u_f(v_i,v_{d,i},t)$}{Adaptive feedback-enhancement term}
\nomenclature[F08]{$\hat{\chi}_i$}{Adaptive variable in the feedback law}
\nomenclature[F09]{$\epsilon_i$}{Positive feedback gain}
\nomenclature[F10]{$\gamma_i$}{Positive adaptive gain}
\nomenclature[F11]{$\delta_i$}{Positive damping coefficient in the adaptation law}
\nomenclature[F12]{$E(t)$}{Robot pose error used in the logarithmic safety term}
\nomenclature[F13]{$O$}{Safety-zone bound in the logarithmic barrier}
\nomenclature[F14]{$\zeta$}{Safety offset}
\nomenclature[F15]{$\bar{V}$}{Lyapunov function}
\nomenclature[F16]{$\dot{\bar{V}}$}{Time derivative of the Lyapunov function}
\nomenclature[F17]{$d_i^{\star}$}{Upper bound on $|d_i|$}
\nomenclature[F18]{$\kappa_i$}{Positive constant used in the Cauchy--Schwarz bound}
\nomenclature[F19]{$\mu$}{Exponential decay-rate constant}
\nomenclature[F20]{$\ell$}{Residual bound in $\dot{\bar{V}}\le -\mu \bar{V}+\ell$}
\nomenclature[F21]{$\rho(t)$}{Normalized safety variable, $\rho(t)=E(t)/O$}
\nomenclature[F22]{$\rho_{\mathrm{warn}}$}{Warning threshold for velocity reduction}
\nomenclature[F23]{$T_s$}{Supervisor sampling time}
\nomenclature[F24]{$\varepsilon_s$}{Safety guard-band margin before $E=O$}
\nomenclature[F25]{$\zeta_{\mathrm{msr}}$}{Upper bound on measured planar pose rate}
\nomenclature[F26]{$\mathcal{S}_{\mathrm{safe}}$}{Supervisor safe set, $\mathcal{S}_{\mathrm{safe}}:=\{(x,y):E(t)<O\}$}
\nomenclature[F27]{$a_v^b$}{Braking linear deceleration limit}
\nomenclature[F28]{$a_\omega^b$}{Braking angular deceleration limit}
\nomenclature[F29]{$\Delta p_{\max}$}{Maximum admissible pose jump}
\nomenclature[F30]{$\Delta \theta_{\max}$}{Maximum admissible heading jump}
\nomenclature[F31]{$T_{\mathrm{lost}}$}{Localization-loss timeout}
\nomenclature[F32]{$d_s$}{Safe-site arrival tolerance}
\nomenclature[F33]{$d_i^\ast$}{Upper bound on the lumped uncertainty $d_i$ for wheel $i$}

\section{Introduction}
Reliable goal-reaching is essential for large-scale mobile robots in harsh, partially known environments such as mining, construction, and forestry \cite{galati2022adaptive}. Accurate and safe target reaching without continuous human intervention improves efficiency, reduces downtime, and minimizes worker exposure to hazardous areas \cite{shahna2025robust}. Consequently, robust goal-reaching and navigation are key enablers for wider adoption of autonomous robots in these industries.

Recently, model-free reinforcement learning (RL) has seen growing use in mobile robotics, especially for tasks that require high performance and efficiency. At the same time, obtaining strict safety guarantees in the implementation of RL remains a major open problem. Many studies on RL with guarantees focus on driving the agent into a designated safe region (goal) or keeping it away from low-reward areas \cite{buhrer2023multiplicative}. For instance, \cite{chow2019lyapunov} proposes embedding Lyapunov functions in a constrained Markov decision process (MDP) framework, combined with an actor–critic update. \cite{yaremenko2024novel} introduces the critic as a Lyapunov function agent, which treats the critic as a Lyapunov surrogate and constrains its updates while leveraging a nominal goal-reaching baseline policy, so that all state–action pairs remain safely explorable. Similarly, \cite{huh2020safe} presents a model-free RL algorithm that merges probabilistic reachability analysis with Lyapunov-based tools to enforce safety, learning during policy evaluation a Lyapunov function that both provides safety guarantees and steers exploration, progressively enlarging the region of states the robot can visit while respecting its constraints, although these guarantees remain only probabilistic. In a related direction, \cite{han2020actor} presents an actor–critic RL scheme with formal stability guarantees for nonlinear, high-dimensional systems by using Lyapunov-based techniques to keep the learned policies stable and enable recovery in the presence of uncertainties. \cite{emam2022safe} merges robust Control Barrier Functions (CBFs) with RL to guide exploration toward high-reward regions in continuous control problems. It embeds a differentiable robust CBF-based safety module inside the soft actor-critic algorithm, so the controller can enforce safety in real time while still improving the navigation policy. In all cases, the robot continues exploring while respecting safety and navigation constraints. However, designing formal safety guarantees using CBFs or Lyapunov-function critics for a large-scale robot with a high-dimensional state space remains challenging. For that reason, despite significant advances in RL for goal-reaching tasks, existing approaches have been developed primarily for light robots, often abstracting away the actuator system, and typically assume that low-level motion actions are handled perfectly in a separate final stage \cite{11458687}. This assumption is not acceptable for large-scale robots, which are actuated by highly nonlinear actuator structures \cite{hyon2019whole, shahna2024integrating, huang2025sequential}. In addition, more components and interfaces mean a higher likelihood of systematic faults than in smaller robots. This risk is further exacerbated by the fact that such systems often operate in harsh environments and in locations that are difficult for humans to access, which complicates inspection and maintenance \cite{11419776}. These factors also make large-scale industries, for example, in mining, more cautious about deploying highly autonomous systems for complex robotic tasks.

To address these challenges, this study presents a hierarchical safety-aware goal-reaching framework for a heavy skid-steered robot with complex in-wheel actuator chains. The framework couples stereo visual localization, constrained RL-based motion-reference generation, learned actuator feedforward compensation, robust adaptive wheel-speed tracking, and supervisory safe-return logic. \textcolor{black}{The paper’s overall contribution is detailed through the following components: 1) Integration of Oriented FAST and Rotated BRIEF simultaneous localization and mapping version 3 (ORB-SLAM3) as the real-time stereo visual pose-estimation interface within the complete control stack of the studied heavy robot; 2) Design of a constrained Q-learning motion planner with acceleration actions, motion constraints, reward shaping, hysteresis, and zero-lock logic for generating smooth and feasible goal-reaching references; 3) Identification of a scaled conjugate-gradient (SCG)-trained deep neural network (DNN) feedforward map from wheel-speed and control-input data, combined with a closed-loop robust adaptive controller (RAC) to compensate for residual actuator uncertainty; 4) Implementation of a sampled supervisory layer that detects safety-boundary violations, localization inconsistencies, and injected faults in the tested scenarios, then switches the robot to braking and safe-return mode; and 5) Experimental validation of the complete framework on asphalt and loose-soil terrain using a 6000 kg robot with complex in-wheel actuator chains. The adopted components, including ORB-SLAM3, Q-learning, SCG-trained DNN regression, and RAC, are established methods. The contribution of this work lies in their problem-specific integration, constraint-aware tuning, actuator-level realization, supervisory safety logic, and experimental validation on a heavy off-road platform. The methodological novelty is therefore limited to the task-specific integration and realization of these components for safe goal-reaching on the studied heavy robot, rather than to the invention of new simultaneous localization and mapping (SLAM), RL, DNN, or RAC algorithms.}

The remainder of the paper is organized as follows. Section 2 presents the vision-based Q-learning-based motion planner for goal-reaching under motion constraints and slippage. Section 3 details the DNN-based actuator feedforward map, feedback RAC, and safety-supervisor logic. Section 4 reports the data collection, testing, physical deployment, and experimental validation of the complete system on the 6000 kg robot. Section 5 discusses scalability and limitations. Section 6 concludes the paper.

\section{Vision-based RL Motion Planning}
As global navigation satellite system (GNSS)-based pose estimation can degrade in large outdoor environments because of propagation effects, stereo visual SLAM is used as the real-time pose source for the proposed control stack\cite{jantos2024aivio, song2022dynavins}. In this work, ORB-SLAM3 introduced in \cite{campos2021accurate} is adopted as an established stereo visual SLAM framework because it provides real-time tracking, loop closing, map fusion, and relocalization capabilities. The localization module's role is to provide the measured robot pose to the motion planner and safety supervisor. The camera pose is written as $T=[R_c,x_{\mathrm{msr}}]\in SE(3)$, where $R_c \in SO(3)$ is the camera rotation matrix and $x_{\mathrm{msr}}=[x,y,\theta]^\top \in \mathbb{R}^3$ is the planar pose used by the goal-reaching controller. The pose $x_{\mathrm{msr}}$ is passed online to the RL motion planner and supervisory safety layer. In the vision-based motion planning module, the problem is modeled as a finite Markov decision process (MDP)
\begin{equation}
\begin{aligned}
\small
\label{35}
\mathcal{M}=(\mathcal{S},\mathcal{A},P,\mathcal{R},\gamma)
\end{aligned}
\end{equation}
where $\mathcal{S}$ is the state space, $\mathcal{A}$ is the action space, $P$ is the state-transition probability kernel, $\mathcal{R}$ is the reward function, and $\gamma \in (0,1)$ is the discount factor for future rewards. The continuous state is given by $x_t$, $y_t$, $\theta_t$, $v_t$, and $\omega_t$ where $(x_t,y_t) \in \mathbb{R}^2$ is the position, $\theta_t$ is the heading, $v_t$ is the linear velocity, and $\omega_t$ is the angular velocity at time $t$. Given a goal position $(x_g, y_g)$, we define

\begin{equation}
\begin{aligned}
\small
\label{35}
d_t &= \sqrt{(x_g - x_t)^2 + (y_g - y_t)^2},\\
e_t &= \operatorname{wrap}_\pi\!\bigl(\operatorname{atan2}(y_g - y_t, x_g - x_t) - \theta_t\bigr)
\end{aligned}
\end{equation}
where $d_t$ is the distance to the goal and $e_t$ is the heading error.
These quantities are discretized into bins, as

\begin{equation}
\label{eq:ocp-cost}
\small
\begin{aligned}
\left\{
\begin{aligned}
&\text{Distance} \quad d^{(1)}, \dots, d^{(N_d)} \quad \text{over} \quad [0, d_{\max}],\\
&\text{Heading error} \quad e^{(1)}, \dots, e^{(N_\theta)} \quad \text{over} \quad [-\pi, \pi),\\
&\text{Linear vel.} \quad v^{(1)}, \dots, v^{(N_v)} \quad \text{over} \quad [v^{\min}, v^{\max}],\\
&\text{Angular vel.} \quad \omega^{(1)}, \dots, \omega^{(N_\omega)} \quad \text{over} \quad [\omega^{\min}, \omega^{\max}]\\
\end{aligned}\right
.
\end{aligned}
\end{equation}
The discrete state is $s_t=(i_d,i_e,i_v,i_\omega)\in\mathcal{S}$ where each index is obtained by binning $(d_t, e_t, v_t, \omega_t)$ and the 4D index is flattened to a single integer. Unlike \cite{yaremenko2024novel}, to generate smooth motion commands, the agent does not act directly on $v$ and $\omega$ but on linear and angular accelerations, as $a_t=(a_{v,t},a_{\omega,t})\in\mathcal{A}$.
The action space $\mathcal{A}$ is a finite grid, as

\begin{equation}
\small
\label{eq:ocp-cost}
\begin{aligned}
\left\{
\begin{aligned}
&a_{v,t} \in \{a_v^{(1)}, \dots, a_v^{(N_{a_v})}\} \subset [a_v^{\min}, a_v^{\max}],\\
&a_{\omega,t} \in \{a_\omega^{(1)}, \dots, a_\omega^{(N_{a_\omega})}\} \subset [a_\omega^{\min}, a_\omega^{\max}]
\end{aligned}\right
.
\end{aligned}
\end{equation}
where $N_{a_v}$ and $N_{a_\omega}$ denote the numbers of discrete
linear- and angular-acceleration levels, respectively. The action space
$A$ is the Cartesian product of these sets. \textcolor{black}{The proposed Q-learning planner is formulated as a discrete observation/state and discrete action-space problem. The continuous robot observations \((d_t,e_t,v_t,\omega_t)\) are quantized into the discrete state \(s_t=(i_d,i_e,i_v,i_\omega)\), while the control actions are selected from the finite acceleration grid \(a_t=(a_{v,t},a_{\omega,t})\).} For the RL motion-planning layer, the underlying continuous-time surrogate model of the skid-steered robot is $\dot{x}=v\cos\theta$,
$\dot{y}=v\sin\theta$, $\dot{\theta}=\omega$, $\dot{v}=a_v$, and $\dot{\omega}=a_\omega$.
Assuming the acceleration inputs $(a_{v,t},a_{\omega,t})$ are held constant over each sampling interval $[t,t+\Delta t)$ by zero-order hold, the RL environment uses the following discrete-time update:

\begin{equation}
\small
\label{eq:ocp-cost}
\begin{aligned}
v_{t+1} &= \mathrm{sat}\!\left(v_t + a_{v,t}\Delta t,\; v_{\min},\, v_{\max}\right),\\
\omega_{t+1} &= \mathrm{sat}\!\left(\omega_t + a_{\omega,t}\Delta t,\; \omega_{\min},\, \omega_{\max}\right),\\
x_{t+1} &= x_t + v_{t+1}\cos(\theta_t)\Delta t,\\
y_{t+1} &= y_t + v_{t+1}\sin(\theta_t)\Delta t,\\
\theta_{t+1} &= \mathrm{wrap}_{\pi}\!\left(\theta_t + \omega_{t+1}\Delta t\right).
\end{aligned}
\end{equation}
where $\operatorname{sat}(\cdot)$ is a saturation operator and $\Delta t$ is the time step. 
Equation (5) is a first-order semi-implicit discrete-time kinematic approximation used for reference generation in the RL layer: the bounded velocities $v_{t+1}$ and $\omega_{t+1}$ are computed first, and the pose is then propagated over one sampling step. 
This approximation keeps the tabular RL environment simple, while actuator nonlinearities, slip effects, and residual tracking errors are handled by the lower-level DNN-based RAC layer during execution.
The workspace is rectangular; if $(x_{t+1},y_{t+1})$ leaves this workspace, the episode terminates with failure. 
The next discrete state $s_{t+1}$ is obtained by binning $(d_{t+1}, e_{t+1}, v_{t+1}, \omega_{t+1})$ as above. 
Therefore, under the surrogate model and fixed action, the transition kernel $P(s' \mid s,a)$ is deterministic. The goal is to learn a policy $\pi(a \mid s)$ that maximizes the expected discounted return

\begin{equation}
\begin{aligned}
\small
\label{35}
J^\pi = \mathbb{E}_\pi\left[\sum_{t=0}^{T_{\mathrm{ep}}}\gamma^t r_t\right]
\end{aligned}
\end{equation}
where $T_{\mathrm{ep}}$ is the episode terminal time when the goal is reached, a timeout occurs, or the agent leaves the workspace. The immediate reward is a shaped cost that combines task progress, smoothness, and stability, as
\begin{equation}
\begin{aligned}
\small
\label{35}
r_t = \mathcal{R}(s_t,a_t,s_{t+1}) = R_{\mathrm{task}} + R_{\mathrm{shape}} 
\end{aligned}
\end{equation}
where $s_t$ and $s_{t+1}$ denote the current and next discrete states, and $a_t$ is the applied action.
The term $R_{\text {task }}\left(s_t, a_t, s_{t+1}\right)$ encodes the primary objective, including:

\begin{itemize}
    \item [1] Step cost: $r_{\text{step}} = -k_{\text{step}}$
    \item [2] Distance progress: $r_d = k_d (d_t - d_{t+1})$
    \item [3] Timeout shaping: $r_{\text{timeout}} = -k_{\text{timeout}} d_T$
\end{itemize}

where $k_{\text {step }}>0$ penalizes long trajectories, $k_d>0$ weights distance reduction, $k_{\text {timeout }}>0$ weights timeout penalties, and $d_T$ is the final distance to the goal at timeout.
Together, these terms define the basic task-level preference for reaching the goal quickly and avoiding timeout far from the goal:
\begin{equation}
\begin{aligned}
\small
\label{35}
R_{\text{task}}= r_{\text{step}}+r_d+r_{\text{timeout}},
\end{aligned}
\end{equation}
The shaping term
$R_{\text{shape}}(s_t,a_t,s_{t+1})$ refines the behavior by encoding
preferences over how the goal is reached, in terms of approach direction,
smoothness, and avoidance of oscillatory motion. The shaping reward is defined as
\begin{equation}
\begin{aligned}
\small
\label{eq:Rshape}
R_{\text{shape}} =&\; r_\theta + r_\omega + r_v^{\parallel} + r_v^{\perp} + r_a
+ r_{\text{hyst}} + r_{\text{goal}} + r_{\mathrm{flip}} \\
&\; + r_{\text{inc}} + r_{\text{stall}} + r_{\text{stop}} + r_{\text{sign}} .
\end{aligned}
\end{equation}

Let $a_{v,t}$ and $a_{\omega,t}$ be the applied linear and angular accelerations, respectively. Define
\begin{equation}
\begin{aligned}
\small
\label{eq:Rshape}
&\Delta e := |e_{t+1}|-|e_t|,
\theta_{\text{stop}} := \frac{\omega_{t+1}^2}{2a_\omega^B},\\
&\text{excess} := \max\!\left(0,\theta_{\text{stop}}-(|e_{t+1}|+e_{\text{pad}})\right).
\end{aligned}
\end{equation}
With $s_0=\operatorname{sign}(e_0)$, define
\begin{equation}
\begin{aligned}
\small
\label{eq:Rshape}
\text{wrong} := \max\!\left(0,-s_0\,\omega_{t+1}-\omega_{\text{db}}\right).
\end{aligned}
\end{equation}
Assume positive parameters $k_\theta$, $k_\omega$, $k_v$, $k_{\text{lat}}$, $k_{a_v}$, $k_{a_\omega}$, $k_{ws}$, $k_d$, $k_{\omega\mathrm{flip}}$, $k_{\text{head,inc}}$, $k_{\text{head,stall}}$, $k_{\omega\text{stop}}$, $k_{\omega\text{sign}}$, together with thresholds $e_{\text{db}}$, $\omega_{\text{db}}$, $\Delta e_{\text{head}}$, $e_{\text{pad}}$, and $d_{\text{goalTol}}$. The reward terms are defined as

\begin{itemize}
    \item[1.] $r_\theta = k_\theta \bigl(|e_t|-|e_{t+1}|\bigr)$
    \item[2.] $r_\omega = -k_\omega \dfrac{1+\cos(|e_{t+1}|)}{2}\,\omega_{t+1}^2$
    \item[3.] $r_v^{\parallel} = k_v\,v_{t+1}\bigl(\max(0,\cos(e_{t+1}))\bigr)^2$
    \item[4.] $r_v^{\perp} = -k_{\text{lat}}\,v_{t+1}^2\sin^2(e_{t+1})$
    \item[5.] $r_a = -k_{a_v}a_{v,t}^2
    -k_{a_\omega}\!\left(0.5+0.5\,\dfrac{1+\cos(|e_{t+1}|)}{2}\right)a_{\omega,t}^2$
    \item[6.] $r_{\text{hyst}} = -k_{ws}\max(0,|\omega_{t+1}|-\omega_{\text{db}})^2,
    \quad \text{if } |e_{t+1}|<e_{\text{db}}$
    \item[7.] $r_{\text{goal}} = k_d\,d_t,
    \quad \text{if } d_{t+1}\le d_{\text{goalTol}}$
\item [8.] $r_{\mathrm{flip}} = -k_{\omega\mathrm{flip}}$ if $\omega_{t+1}$ changes sign and $|e_{t+1}| < e_{\text{db}}$
\item [9.] $r_{\text{inc}} = -k_{\text{head,inc}} \max(0,\Delta e)$
\item [10.] $r_{\text{stall}} = -k_{\text{head,stall}} |e_{t+1}|$ if $|\Delta e| < \Delta e_{\text{head}}$ and $|e_{t+1}| > 0$
\item [11.] $r_{\text{stop}} = -k_{\omega\text{stop}}\,\text{excess}^2$
\item [12.] $r_{\text{sign}} = -k_{\omega\text{sign}}\,\text{wrong}^2$ if $\text{wrong} > 0$
\end{itemize}
Reward terms with stated conditions are set to zero when their conditions are not satisfied; the timeout penalty is applied only at timeout.
The reward terms jointly promote distance reduction, heading-error correction, smooth acceleration, forward motion when aligned with the goal, and suppression of oscillatory steering near the goal. The hysteresis, sign-flip, predictive stopping, and wrong-sign angular penalties are included to reduce residual heading oscillations and generate smoother motion references for the heavy skid-steered platform. The distance-progress and heading-progress terms can be interpreted as decreasing a Lyapunov-like potential $\Phi\left(s_t\right)=k_d d_t+k_\theta\left|e_t\right|$, since $r_d+r_\theta=\Phi\left(s_t\right)-\Phi\left(s_{t+1}\right)$. Thus, these terms guide the policy toward smaller distance and heading error. The remaining shaping terms are not potential-based in the strict sense; they intentionally modify the control objective to favor smooth, non-oscillatory, and monotone motion references suitable for the heavy skid-steered platform.
The agent maintains a tabular action value function
$Q:\mathcal{S}\times\mathcal{A}\rightarrow\mathbb{R}$ initialized to zero.
At each time step, a temporal-difference (TD) update is applied
\begin{equation}
\small
\label{eq:Q-update}
Q(s_t, a_t)
\leftarrow
Q(s_t, a_t) + \alpha\, \delta_t,
\end{equation}
where $\alpha > 0$ is the learning rate and $\delta_t$ is the TD error.
The TD error is given by

  \begin{equation}
  \small
  \label{eq:delta-Qlearning}
  \delta_t
  = r_t + \gamma \max_{a' \in \mathcal{A}} Q(s_{t+1}, a') - Q(s_t, a_t),
  \end{equation}

where $r_t = R(s_t, a_t, s_{t+1})$ is the immediate reward. In this work, Q-learning was implemented and trained under the same state discretization, reward shaping, exploration schedule, and action constraints. During training, actions are drawn from an $\varepsilon$-greedy policy 
with respect to $Q$:
\begin{equation}
\small
a_t =
\begin{cases}
\text{random action} & \text{with probability } \varepsilon_t, \\
\displaystyle \arg\max_{a \in \mathcal{A}} Q(s_t, a)
& \text{with probability } 1 - \varepsilon_t,
\end{cases}
\end{equation}
where the exploration parameter $\varepsilon_t$ is decayed exponentially from $\varepsilon_0$ to $\varepsilon_{\text {final }}$ over the training episodes, while a purely greedy policy ( $\varepsilon_t=0$ ) is used at evaluation. In addition to reward shaping, two policy-side mechanisms act directly on the angular dynamics near the desired heading and goal. When the agent is well aligned and rotates slowly ( $\left|e_t\right|<e_{\mathrm{db}},\left|\omega_t\right|<\omega_{\mathrm{db}}$ ), the angular acceleration is clamped to $a_{\omega, t}=0$, which suppresses small oscillations. In a small goal neighborhood ( $\left|e_t\right| \leq e_{\text {lock }}, d_t \leq d_{\text {lock }}$ ), further constraints are applied: during training, a bounded braking command drives $\omega_t$ toward zero within acceleration limits, and during evaluation a hard clamp enforces $\omega_{t+1}=0$ while $v_t$ still follows the selected $a_{v, t}$. This corresponds to an MDP with state-dependent action constraints and modified transitions near the goal, used to stabilize the policy and eliminate residual heading wiggles.

\begin{figure}[h!]
\hspace*{-0.0cm} 
\centering
\scalebox{0.85}{\includegraphics[trim={0cm 0.0cm 0.0cm 0cm},clip,width=\columnwidth]{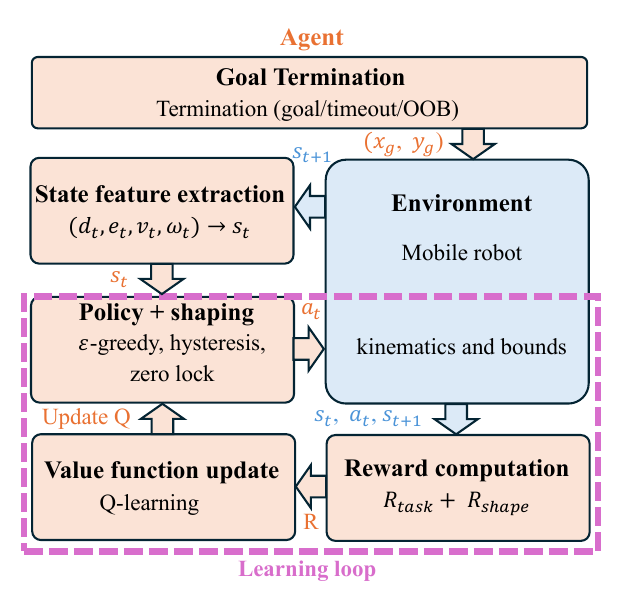}}
\caption{Architecture of the RL motion planning.}
\label{asdasdasdgoial}
\end{figure}

\begin{figure*}[h!]
\hspace*{-0.0cm} 
\centering
\includegraphics[width=0.95\textwidth, height=6.5cm]{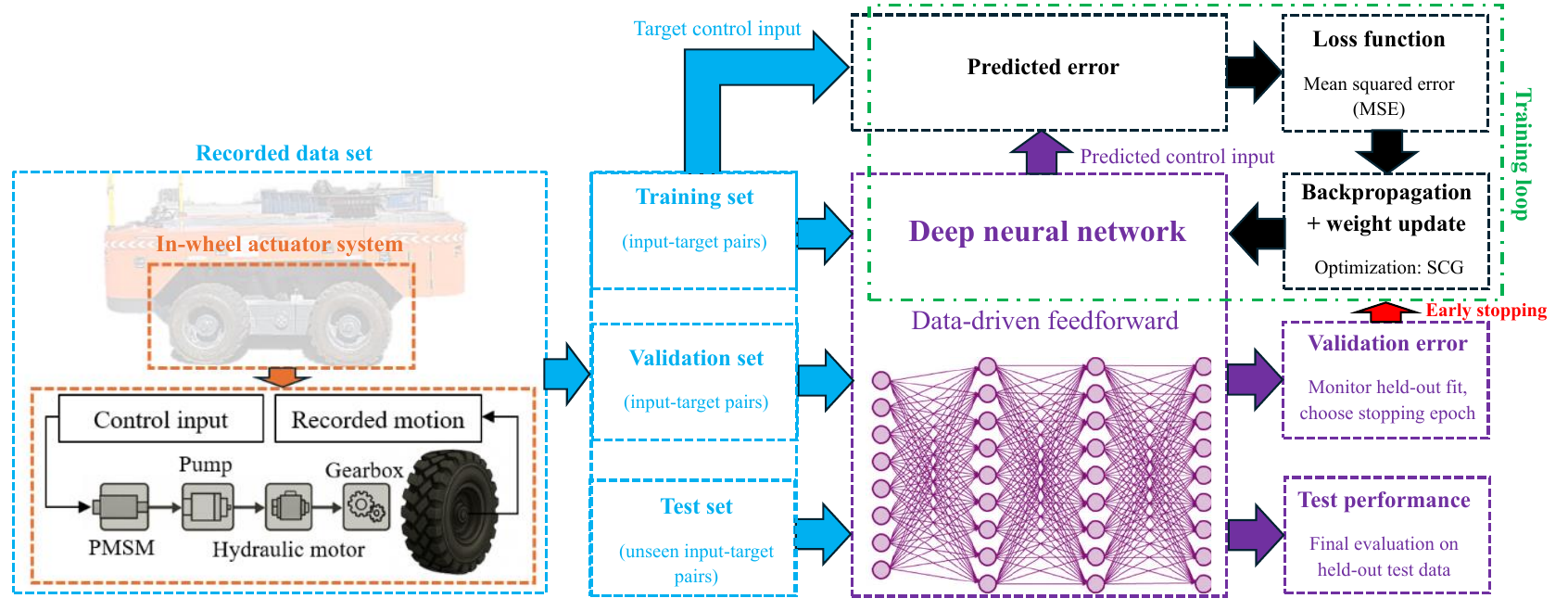}
\caption{Training procedure for the SCG-trained DNN actuator feedforward map using data collected from the studied in-wheel actuator chain. \textcolor{black}{The actuator-chain block is a schematic representation of the experimental platform, not a general model claimed for all large robots \cite{11458687, 11419776}}.}
\label{circcccusasddlar}
\end{figure*}

Fig. \ref{asdasdasdgoial} shows the closed-loop RL motion planner, where the environment updates the continuous state from the applied acceleration and goal, which is then discretized into a Markov state. This state is fed to an $\varepsilon$-greedy policy with hysteresis and zero-lock shaping to generate the next action. In parallel, each transition and its reward are used to update the Q-table using Q-learning.

\section{DNN-based RAC for In-wheel Actuator Mechanism}

\subsection{SCG-trained DNN Actuator Feedforward Map}
\textcolor{black}{For each wheel of the studied robot, the control input of the $i$th wheel $u_i$ is applied to the actuator system chain, which consists of the commanded drive unit, hydraulic motor, gearbox, and wheel-speed response, as schematically illustrated in Fig. \ref{circcccusasddlar}.} If $r$ is defined as the radius of the wheels, for each applied input, the resulting tangential (linear) wheel speed at the rim $v_i = r \omega_i$ is measured and recorded, providing input–output datasets for subsequent data-driven actuator feedforward mapping. After collecting the recorded actuator data
$\{v_i(k), u_i(k)\}_{k=1}^{N_{\mathrm{data}}}$, where
$N_{\mathrm{data}}$ is the number of recorded actuator samples, \textcolor{black}{the problem is formulated as supervised regression, where a DNN approximates a static feedforward map from wheel speed to nominal control input}, $\hat{u}_i = f_{\beta_i}(v_i)$, with $\beta$ collecting all weights and biases. \textcolor{black}{Because the map uses wheel speed as its input, it is a quasi-static nominal compensation term rather than a full inverse actuator-dynamics model. Acceleration-dependent dynamics, delays, hysteresis, terrain memory, and operating-envelope mismatch are treated as residual uncertainties by the feedback RAC.} The data are split into three disjoint sets: a training set used to adjust $\beta$, a validation set used to monitor held-out fitting performance and determine stopping, and a test set used only for final evaluation.
The model is a fully connected feedforward network with one scalar input,
$L \in \mathbb{N}$ hidden layers, where hidden layer $j$ contains
$n_j \in \mathbb{N}$ neurons, and one scalar output. Denoting the input as $a^{(0)}=v_i$, the hidden layers as $a^{(1)}, a^{(2)}, \ldots, a^{(L)}$, and the output as $\hat{u}_i=a^{(L+1)}$, the forward pass for one sample can be written as
\begin{equation}
\small
\label{eq:pot-shaping-Rtilde}
\begin{array}{cc}
z^{(1)}=W^{(1)} a^{(0)}+b^{(1)}, & a^{(1)}=\phi\left(z^{(1)}\right), \\
z^{(2)}=W^{(2)} a^{(1)}+b^{(2)}, & a^{(2)}=\phi\left(z^{(2)}\right), \\
\vdots \\
z^{(L+1)}=W^{(L+1)} a^{(L)}+b^{(L+1)}, & a^{(L+1)}=\hat{u}_i,
\end{array}
\end{equation}

where $W^{(j)}$ and $b^{(j)}$ are the weight matrices and bias vectors of layer $j$, and $\phi(\cdot)$ is a nonlinear activation function applied elementwise in each hidden layer (for example, a sigmoid or ReLU). Over the training set $\left\{\left(v_i^{(k)}, u_i^{(k)}\right)\right\}_{k=1}^{N_{\text {train }}}$ where $N_{\text {train }}$ is the number of training samples, the network parameters are learned by minimizing the mean squared error cost function

\begin{equation}
\small
\begin{aligned}
\label{eq:pot-shaping-Rtilde}
J(\beta_i)=
\frac{1}{N_{\mathrm{train}}}
\sum_{k=1}^{N_{\mathrm{train}}}
\left(
u_i^{(k)}-f_{\beta_i}(v_i^{(k)})
\right)^2 .
\end{aligned}
\end{equation}

The network parameters are trained using backpropagation with the scaled conjugate-gradient algorithm. The validation loss is monitored during training, and the parameter set with the minimum validation error is selected by early stopping. The final frozen network is then evaluated once on the held-out test set to estimate feedforward performance within the recorded actuator operating envelope.
The training procedure for the DNN actuator feedforward map is shown in Fig. \ref{circcccusasddlar}. \textcolor{black}{After training and validation, the frozen network is used as a nominal actuator feedforward map,}
\begin{equation}
\begin{aligned}
\small
\label{3adasd5}
\textcolor{black}{u_{FF,i}(v_{d,i}) = f_{\beta_i^\star}(v_{d,i}),}
\end{aligned}
\end{equation}
where $v_{d,i}$ is the desired wheel speed generated from the RL
motion-planning layer, and $\beta_i^\star$ denotes the selected trained
DNN parameter vector for wheel $i$.

\textit{\textcolor{black}{Remark 3.1}}. The scalar feedforward map is used as a nominal compensation term, not as a terrain-aware temporal actuator model. Since reliable labeled inputs for terrain class, slip ratio, payload variation, and input history were not available in the present experiments, these effects are handled as residual uncertainties. Sequence models such as long short-term memory (LSTM) or gated recurrent unit (GRU) networks \cite{Alcayaga2025} are left for future work when richer terrain-dependent data are available.

\begin{figure*}[h!]
\hspace*{-0.0cm} 
\centering
\includegraphics[width=0.85\textwidth, height=9.5cm]{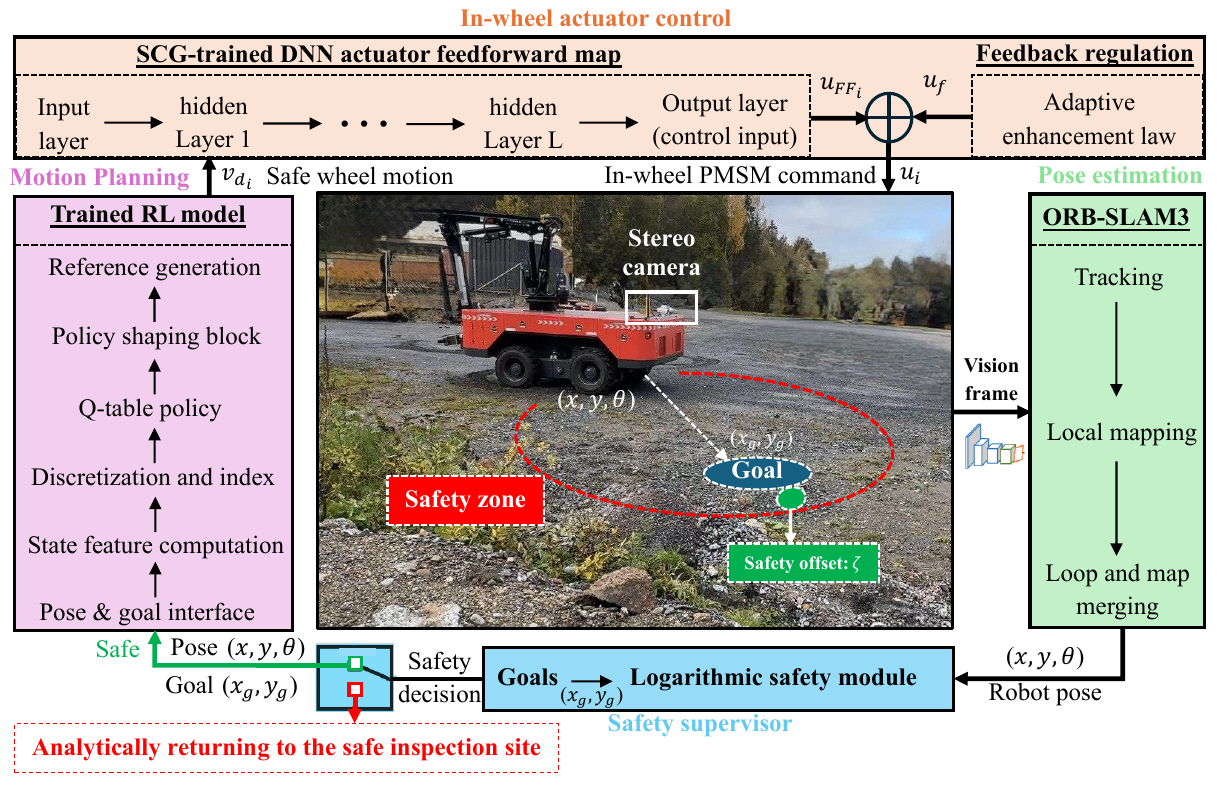}
\caption{The proposed framework in real time.}
\label{ADADGSFHG}
\end{figure*}

\subsection{Feedback RAC Enhancement}
When the robot encounters operating conditions not fully represented by the recorded data or external disturbances, the nominal feedforward map alone may not provide sufficient tracking accuracy. In these cases, using the DNN-based feedforward map leads to wheel velocities that differ from the desired values, which results in a nonzero tracking error. The wheel-speed tracking error is defined as $e_i(t)=v_i-v_{d_i}$, where $v_i$ is the instantaneous wheel speed measurement. We represent the actuator-level tracking error dynamics in input-error form as
\begin{equation}
\begin{aligned}
\small
\label{3asdas5}
A_i \dot e_i =
u_i(t)-u_{FF,i}(v_{d,i})+d_i(v_{d,i},e_i,t),
\end{aligned}
\end{equation}
where \(A_i>0\) is an unknown control coefficient and \(d_i(\cdot)\) denotes the lumped residual uncertainty. This term includes the approximation error of the static feedforward map, slip-induced disturbances, acceleration-dependent effects, actuator delays, hysteresis, and terrain-dependent variations. Therefore,

\begin{equation}
\begin{aligned}
\small
\label{35adasd}
\dot e_i =
A_i^{-1}
\left[
u_i(t)-u_{FF,i}(v_{d,i})+d_i(v_{d,i},e_i,t)
\right].
\end{aligned}
\end{equation}

\textit{Assumption 3.1.} The actuator coefficient satisfies
$0 < A_{i,\min} \leq A_i \leq A_{i,\max}$, where $A_{i,\min}$ and
$A_{i,\max}$ are positive lower and upper bounds on $A_i$, respectively.
The lumped residual uncertainty $d_i(\cdot)$ is bounded and locally
Lipschitz continuous \cite{11419776, tan2025fixed}.

The control input is defined as

\begin{equation}
\begin{aligned}
\small
\label{vbfvfvf39}
u_{i} (t) =& u_{\text{FF}_i} ({v}_{d_i}) +u_f(v_i, v_{d_i}, t)\\
=& u_{\text{FF}_i} ({v}_{d_i}) -  \frac{1}{2} \epsilon_i e_i - \gamma_i e_i \log^2 (\frac{O}{O-E}) \hat{\chi}_i
\end{aligned}
\end{equation}
where $\gamma_i \in \mathbb{R}_{+}$ and $\epsilon_i \in \mathbb{R}_{+}$ are positive constants, $u_f(.)$ is the feedback control enhancement, and $\hat{\chi}_i$ is the implemented adaptive law, defined as
\begin{equation}
\begin{aligned}
\small
\label{3aadasdasd2}
\dot{\hat{\chi}}_i (t) = -\frac{1}{2}\delta_i \hat{\chi}_i+\gamma_i e^2_i \log^2 (\frac{O}{O-E})
\end{aligned}    
\end{equation}
where $E(t)$ is a strictly positive function satisfying $E(t)<O$, with $O \in \mathbb{R}_{+}$a positive constant, $\delta_i \in \mathbb{R}_{+}$, and $\hat{\chi}_i\left(t_0\right) \in \mathbb{R}_{+}$. Following \cite{SHAHNA2025106516}, and as illustrated in Eqs. \eqref{vbfvfvf39} and \eqref{3aadasdasd2}, a logarithmic barrier function is incorporated in the proposed control scheme as the safety component for the overall system. Without loss of generality, we reset the initial robot position to $(0,0)$ regardless of its heading angle, and let $E(t)$ and $O$ denote the robot pose error, and the goal pose error (with safety offset), respectively, defined as 

\begin{equation}
\small
\label{eq:ocp-cost}
\begin{aligned}
\left\{
\begin{aligned}
&\text{Current Robot Zone:} \hspace{0.05cm} {E}(t) = \sqrt{{(x - \frac{{x}_g}{2}})^2 + {(y - \frac{{y}_g}{2}})^2},\\
&\text{Safety Robot Zone:} \hspace{0.05cm} O = \zeta + \sqrt{(\frac{{x}_g}{2})^2 + (\frac{{y}_g}{2})^2}
\end{aligned}\right
.
\end{aligned}
\end{equation}
$x$ and $y$ denote the robot's real-time position, and \(O\) defines a safety circle centered at the midpoint between the initial position \((0,0)\) and the goal \((x_g,y_g)\), with radius equal to one-half of the start-to-goal distance plus the safety offset \(\zeta\). If $E(t)$ reaches the safety boundary, numerical singularities occur, and the execution of the unsafe control operation is terminated \cite{SHAHNA2025106516}. At this time, a safety supervisor with a well-defined state machine limits velocity near the barrier, applies a deterministic braking profile once limits are violated, and latches the system into a predefined numerical motion planning and control mode that returns the robot to a safe site with respect to its current position. When the robot reaches the goal and needs to move to a new one, the SLAM pose and goal are updated. The built framework for implementation in real time is shown in Fig. \ref{ADADGSFHG}.

\subsection{\textcolor{black}{Safety Supervisor Logic and Boundedness Analysis}}

The logarithmic safety mechanism is implemented together with a sampled supervisory state machine. The supervisor is not intended to provide a general proof of full-system safety under arbitrary localization or terrain faults, but to enforce a deterministic response when the monitored safety variable approaches its admissible boundary or when a localization inconsistency is detected. The supervisor has four modes: \emph{Run}, \emph{Warning}, \emph{Brake}, and \emph{Safe-return}. In the \emph{Run} mode, the RL planner provides the active goal-reaching reference. In the \emph{Warning} mode, the admissible reference velocity is reduced. In the \emph{Brake} mode, the desired linear and angular velocities are ramped toward zero using bounded deceleration. After braking, the supervisor latches into \emph{Safe-return} mode and replaces the current goal by the predefined inspection site. The supervisor is evaluated every $T_s>0$ seconds. Let
$\rho(t)=\frac{E(t)}{O}$
denote the normalized safety variable. The system remains in \emph{Run} mode while $\rho(t_k)<\rho_{\mathrm{warn}}$ and no localization inconsistency is detected. If $\rho(t_k)\geq \rho_{\mathrm{warn}}$, the system enters \emph{Warning} mode and scales down the commanded velocity. If
$E(t_k)\geq O-\varepsilon_s$,
where $\varepsilon_s>0$ is the safety guard-band margin, or if a localization fault is detected, the supervisor enters \emph{Brake} mode. Localization faults are detected from abnormal pose increments, abnormal heading jumps, or loss of the localization stream for longer than a prescribed time. The braking command is generated as

\begin{equation}
\begin{aligned}
\small
\label{adasdghgfg}
v_{b,d,k+1}=&\mathrm{sat}\left(v_{b,d,k}-a_v^bT_s,0,v_{b,d,k}\right),
\\
\omega_{b,d,k+1}=&\mathrm{sign}(\omega_{b,d,k})
\max\left(0,|\omega_{b,d,k}|-a_\omega^bT_s\right),
\end{aligned}
\end{equation}

where $v_{b,d,k}$ and $\omega_{b,d,k}$ are the desired body linear and
angular velocities at supervisor sample $k$, and $a_v^b > 0$ and
$a_\omega^b > 0$ are the deterministic braking limits. The supervisor is latched after a brake or localization-fault event; therefore, normal goal-reaching is not automatically resumed during the experiment. We next provide the boundedness argument for the guard-band part of the supervisor. Define the safe set
\begin{equation}
\begin{aligned}
\small
\label{asdsbvgfnfgfg}
\mathcal{S}_{\mathrm{safe}}
:=
\{(x,y):E(t)<O\}.
\end{aligned}
\end{equation}

During each active goal-reaching segment, the center of the safety circle is fixed. Assume that the measured planar pose is absolutely continuous and that its rate is bounded as

\begin{equation}
\begin{aligned}
\small
\label{asdsasfgbvgfnfgfg}
\|\dot{x}_{\mathrm{msr}}(t)\|_2 \leq \zeta_{\mathrm{msr}}
\end{aligned}
\end{equation}

where $x_{\mathrm{msr}}=[x,y]^\top$. Since $E(t)$ is the Euclidean distance between the measured robot position and the center of the safety circle, it follows that, for any $t\in[t_k,t_{k+1}]$,

\begin{equation}
\begin{aligned}
\small
\label{asdsasfgbasdasvgfnfgfg}
&E(t)-E(t_k)\\
&\leq
\|x_{\mathrm{msr}}(t)-x_{\mathrm{msr}}(t_k)\|_2
\leq
\int_{t_k}^{t}\|\dot{x}_{\mathrm{msr}}(\tau)\|_2 d\tau
\leq
\zeta_{\mathrm{msr}}(t-t_k).
\end{aligned}
\end{equation}
Therefore,
\begin{equation}
\begin{aligned}
\small
\label{asdsassadfgbasdasvgfnfgfg}
E(t)\leq E(t_k)+\zeta_{\mathrm{msr}}T_s,
\qquad t\in[t_k,t_{k+1}].
\end{aligned}
\end{equation}
If the guard band satisfies
\begin{equation}
\begin{aligned}
\small
\label{asdsassjfg}
\varepsilon_s>\zeta_{\mathrm{msr}}T_s,
\end{aligned}
\end{equation}
then whenever $E(t_k)\leq O-\varepsilon_s$, we have
\begin{equation}
\begin{aligned}
\small
\label{asdsassjfkhg}
E(t)\leq O-\varepsilon_s+\zeta_{\mathrm{msr}}T_s<O
\end{aligned}
\end{equation}
for all $t\in[t_k,t_{k+1}]$. Hence, under the measured-pose-rate bound and sampling assumptions, the monitored safety variable remains below the logarithmic boundary between two supervisor evaluations.

\subsection{Stability Analysis}
Define the following Lyapunov function for the four interactive wheels within the whole robot system.
\begin{equation}
\begin{aligned}
\small
\label{35asdasfadfsdgfsg}
\bar{V}=\sum_{i=1}^{4}\left(\frac{A_i}{2}e_i^2+\frac{1}{2}\hat{\chi}_i^2\right).
\end{aligned}
\end{equation}
Taking the derivative of \eqref{35asdasfadfsdgfsg} and using \eqref{35adasd}, we have
\begin{equation}
\begin{aligned}
\small
\label{37}
\dot{\bar{V}}= & \sum_{i=1}^4 e_i [u_i - u_{\text{FF}_i} + d_i]+ \hat{\chi}_i \dot{\hat{\chi}}_i
\end{aligned}
\end{equation}
Substituting \eqref{vbfvfvf39}, we get
\begin{equation}
\begin{aligned}
\small
\label{37}
\dot{\bar{V}}= & \sum_{i=1}^4 -\frac{1}{2} \epsilon_i e^2_i - \gamma_i e^2_i \log^2 (\frac{O}{O-E}) \hat{\chi}_i + e_i d_i+ \hat{\chi}_i \dot{\hat{\chi}}_i
\end{aligned}
\end{equation}
In view of \textit{Assumption 3.1}, introduce an unknown nonnegative constant $d_i^*$ satisfying $\left|d_i\right| \leq d_i^*$. By substituting \eqref{3aadasdasd2},
\begin{equation}
\begin{aligned}
\small
\label{37}
\dot{\bar{V}} \leq & \sum_{i=1}^4 - \frac{1}{2} \epsilon_i e^2_i - {\gamma_i e^2_i \log^2 (\frac{O}{O-E}) \hat{\chi}_i} + |e_i| \hspace{0.15cm} d^*_i\\
&-\frac{1}{2}\delta_i \hat{\chi}^2_i + {\gamma_i e^2_i \log^2 (\frac{O}{O-E}) \hat{\chi}_i}
\end{aligned}
\end{equation}
Using the Cauchy–Schwarz inequality, we obtain
\begin{equation}
\begin{aligned}
\small
\label{37}
\dot{\bar{V}} \leq & \sum_{i=1}^4 - \frac{1}{2} \epsilon_i e^2_i + \frac{1}{2} \kappa_i e^2_i +  \frac{1}{2\kappa_i} {d^*_i}{^2} -\frac{1}{2}\delta_i \hat{\chi}^2_i
\end{aligned}
\end{equation}
with $\epsilon_i, \kappa_i>0$ chosen so that $\epsilon_i$ exceeds $\kappa_i$. Hence,
\begin{equation}
\begin{aligned}
\small
\label{3afsdfsdf7}
\dot{\bar{V}} \leq & - \sum_{i=1}^4 \frac{1}{2} (\epsilon_i - \kappa_i) e^2_i  +  \frac{1}{2\kappa_i} {d^*_i}{^2} -\frac{1}{2}\delta_i \hat{\chi}^2_i
\end{aligned}
\end{equation}
From \eqref{35asdasfadfsdgfsg} and \eqref{3afsdfsdf7}, we have $\dot{\bar{V}} \leq - \mu {\bar{V}} + \ell $ where
\begin{equation}
\begin{aligned}
\small
\label{37}
\mu:=\min _{i=1, \ldots, 4} \min \left\{A_i^{-1}\left(\epsilon_i-\kappa_i\right), \delta_i\right\}, \quad \ell=\sum_{i=1}^{4}\frac{(d_i^\ast)^2}{2\kappa_i}.
\end{aligned}
\end{equation}
\textcolor{black}{It follows that, in the sense of Definition 1 in \cite{shahna2025robudfdsfst}, and under Assumption 3.1, the closed-loop actuator-level wheel-speed tracking subsystems are uniformly ultimately bounded, with exponential convergence to a compact residual set whose size depends on the disturbance bound.}

\begin{table}[t]
\centering
\caption{\textcolor{black}{Computational environment for offline training, testing, and validation.}}
\label{tab:computing_env}
\resizebox{\columnwidth}{!}{%
\begin{tabular}{ll}
\hline
Item & Specification \\
\hline
Computer model & Dell XPS 15 9530 \\
Operating system & Windows 64-bit, x64-based processor \\
Processor & 13th Gen Intel Core i7-13700H, 2.40 GHz \\
Installed RAM & 32 GB, 31.7 GB usable \\
Graphics memory & 8 GB, multiple GPUs installed \\
Storage & 954 GB \\
RL training/evaluation & MATLAB offline simulation \\
DNN training & MATLAB Neural Network Toolbox \\
Physical validation & 6000 kg robot, Beckhoff PC, cameras, sensors \\
Online deployment & Frozen Q-table and frozen DNN map\\
MATLAB/Simulink version & R2024b\\
\hline
\end{tabular}%
}
\end{table}

\begin{table*}[t]
\centering
\caption{\textcolor{black}{Repeated-run simulation statistics for the proposed RL motion planner over $N=30$ independent runs}.}
\label{tab:repeated_run_stats}
\small
\begin{tabular}{lccc}
\hline
Metric & Mean $\pm$ Std. Dev. & Variance & 95\% Confidence Interval \\
\hline
Position RMSE (m)              & $0.0350 \pm 0.0048$ & $2.30 \times 10^{-5}$ & $[0.0332,\; 0.0368]$ \\
Time-to-goal (s)               & $55.0 \pm 4.6$      & $21.16$               & $[53.3,\; 56.7]$ \\
Heading sign flips             & $0.60 \pm 0.50$     & $0.25$                & $[0.41,\; 0.79]$ \\
$\mathrm{TV}(\omega)$ (rad/s)  & $0.85 \pm 0.09$     & $8.10 \times 10^{-3}$ & $[0.82,\; 0.88]$ \\
Near-goal oscillation index    & $0.030 \pm 0.010$   & $1.00 \times 10^{-4}$ & $[0.026,\; 0.034]$ \\
\hline
\end{tabular}
\end{table*}

\section{Experimental Validation}
\textcolor{black}{The experiments validate the integrated framework under the reported operating conditions, conservative velocity and acceleration limits, and specified fault-injection scenarios.} The platform is a 6000 kg skid-steered off-road robot equipped with stereo RGB cameras and wheel-speed sensing, with data logged through a Beckhoff PC. \textcolor{black}{All learning stages were performed offline and the computational environment used is summarized in Table~\ref{tab:computing_env}. The Q-learning planner was trained in MATLAB simulation using the parameters in Table~\ref{tab:ql_constraints}, and the resulting Q-table was frozen for greedy online action selection. The DNN actuator feedforward maps were trained offline from recorded wheel-speed and control-input data using separate training, validation, and test subsets. Physical experiments were used to validate the frozen learning components inside the complete control stack.}

\begin{table}[h]
\centering
\caption{Constraints and parameters for Q-learning policy.}
\small
\begin{tabular}{lll}
\hline
\textbf{Parameter} & \textbf{Value} & \textbf{Description} \\
\hline
$dt$ & 0.05 & Sim. time step (s) \\
episodes & 30000 & Train. episodes \\
evalEpisodes & 1000 & Greedy eval. episodes \\
goalTol & 0.10 & Goal tol. radius (m) \\
startMinDist & 0.20 & Min. start dis. to goal (m) \\
$[x,y]$ bounds & $[-25,25]$ & Workspace bounds (m) \\
$[\Delta x, \Delta y]$ & [1.0, 1.0] & Grid resolution (m) \\
$n_\theta$ & 24 & Head. bins \\
$[N_v, N_\omega]$ & [4, 5] & Vel. bins \\
$[v_{\min}, v_{\max}]$ & $[0.0, 0.25]$ & Lin. speed limits (m/s) \\
$[\omega_{\min}, \omega_{\max}]$ & $[-0.15, 0.15]$ & Ang. speed limits (rad/s) \\
$[a_{v,\min}, a_{v,\max}]$ & $[-0.10, 0.10]$ & Lin. acc. limits (m/s$^2$) \\
$[a_{\omega,\min}, a_{\omega,\max}]$ & $[-0.02, 0.02]$ & Ang. acc. limits (rad/s$^2$) \\
$[\Delta a_v, \Delta a_\omega]$ & [0.10, 0.02] & Acc. grid step [m/s$^2$, rad/s$^2$] \\
$e_{\text{db}}$ & 0.01 & Hys. head. deadband (rad) \\
$\omega_{\text{db}}$ & 0.001 & Hys. ang. deadband (rad/s) \\
$k_{ws}$ & 1.2 & Hys. spin penalty gain \\
$e_{\text{lock}}$ & 0.03 & Zero lock head. window (rad) \\
$d_{\text{lock}}$ & 0.30 & Zero lock dis. window (m) \\
$\alpha$ & 0.10 & Learning rate (def) \\
$\gamma$ & 0.95 & Discount factor (def) \\
$[\varepsilon_0,\varepsilon_{\text{final}}]$ & $[1.0,10^{-3}]$ & Exp. rate range \\
\hline
\end{tabular}
\label{tab:ql_constraints}
\end{table}

\begin{figure}[h!]
\hspace*{-0.0cm} 
\centering
\scalebox{0.9}{\includegraphics[trim={0cm 0.0cm 0.0cm 0cm},clip,width=\columnwidth]{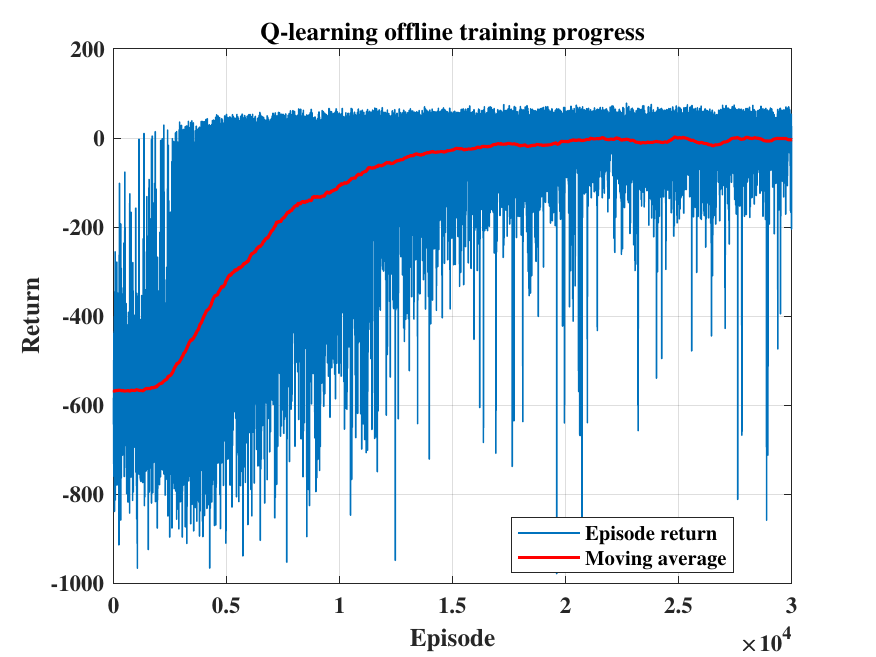}}
\caption{\textcolor{black}{Offline Q-learning motion-planner training progress over 30000 simulation episodes. The raw episode return is shown together with a moving-average curve.}}
\label{sdasfgnk}
\end{figure}

The agent's reward is a weighted sum of distance, heading, velocity, and shaping terms, with weights $k_d=5.0$, $k_\theta=6.0$, $k_v=0.08$, $k_{\text {lat }}=0.9$, $k_\omega=0.28$, $k_{a_v}=0.10$, $k_{a_\omega}=0.10$, $k_{\text {step }}=0.04$, $k_{\text {timeout }}=3.0$, $k_{\omega \text { flip }}=0.85$, $k_{\text {head,inc }}=0.25$, $k_{\text {head,stall }}=0.03$, $\Delta e_{\text {head }}=0.02$, $k_{\omega \text { stop }}= 1.5$, $e_{\text {pad }}=0.01$, and $k_{\omega \text { sign }}=0.8$. As observed in Fig. \ref{ADADGSFHG}, in online operation, if the logarithmic safety supervision allows and $E(t)<O$, the trained RL model receives the current pose $(x,y, \theta)$ from the SLAM module and the goal $(x_g, y_g)$, computes the features $\left(d_t, e_t, v_t, \omega_t\right)$, discretizes them to a Markov state $s_t$, selects an action greedily from the learned Q-table, refines it via hysteresis and zero-lock shaping, and outputs the reference motion $v_d$ for the in-wheel actuator control.
Table \ref{tab:repeated_run_stats} reports the repeated-run simulation statistics of the proposed RL motion planner over multiple independent trials. Here, $TV(\omega)$ denotes the total variation of the angular velocity. The mean, standard deviation, variance, and $95\%$ confidence interval are presented to evaluate the consistency and repeatability of the simulated motion-planning performance. The relatively small spread in the results indicates stable tracking performance and repeatable convergence behavior, showing that the controller performs reliably across different simulation runs. \textcolor{black}{Fig.~\ref{sdasfgnk} shows the offline Q-learning training progress over 30000 simulation episodes. The raw episode return remains noisy because of $\epsilon$-greedy exploration, randomized initial conditions, and penalty-based reward terms. However, the moving-average return increases from highly negative values and approaches a stable plateau near the end of training. The reward function includes step, acceleration, angular-motion, heading-oscillation, and timeout penalties. The observed trend indicates that the learned policy progressively reduces accumulated penalties before the Q-table is frozen and used for greedy online deployment.}

\subsection{Data Collection and Dataset Preparation}
The actuator dataset was collected over approximately 3.5 km of robot operation on asphalt and soft soil, with each terrain contributing 50\% of the samples. The raw dataset contains 1,000,000 paired wheel-speed and control-input samples. Data collection was performed under conservative motion limits because of the size and mass of the platform: linear and angular velocities were limited to 0.35 m/s and 0.15 rad/s, respectively, and linear and angular accelerations to 0.1 m/s$^2$ and 0.02 rad/s$^2$, respectively. Within this safety envelope, command profiles were repeatedly swept from rest to the maximum admissible values to excite the drivetrain across its operating range and to capture variations associated with wheel-terrain interaction. In addition, to reduce noise sensitivity in the safety-layer switching logic, all sensor-derived signals, including stereo visual SLAM pose estimates and hydraulic motor speed measurements, were processed using a first-order low-pass filter. Raw velocity and rpm streams were then aligned, cleaned of obvious spurious spikes, and split into training, validation, and test subsets before normalization.

\begin{table}[ht]
\centering
\caption{DNN configuration and training parameters.}
\small
\begin{tabular}{ll}
\hline
\textbf{Parameter} & \textbf{Value} \\
\hline
Input data          & \texttt{v\_i} \\
Target data         & \texttt{u\_i} \\
Network type            & \texttt{feedforwardnet} \\
Hidden sizes            & {[}320, 210, 105{]} \\
Training function       & \texttt{'trainscg'} \\
Input processing        & \texttt{mapminmax} \\
Output processing       & \texttt{mapminmax} \\
Train ratio             & 0.34 \\
Validation ratio        & 0.33 \\
Test ratio              & 0.33 \\
Goal                    & $1\times 10^{-6}$ \\
Min gradient            & $1\times 10^{-10}$ \\
Max epochs              & 500 \\
Generated function name & \texttt{myTrainedNetFunction} \\
\hline
\end{tabular}
\label{dnnopar}
\end{table}

In accordance with the ratios reported in Table \ref{dnnopar}, the dataset was partitioned into $340{,}000$ training samples, $330{,}000$ validation samples, and $330{,}000$ test samples. The split preserved the $50/50$ terrain balance in each subset. To reduce temporal leakage, the partitioning was performed on contiguous recorded segments rather than by fully random sample-wise shuffling. Four wheel-specific feedforward maps were trained independently using the same network architecture and SCG optimization settings. Model selection was based on the minimum validation loss, and the final reported performance corresponds to the frozen network evaluated on the held-out test set. 
\textcolor{black}{The entries in Table \ref{dnnopar} refer to the MATLAB neural-network implementation used for the actuator feedforward map. The training function \texttt{trainscg} denotes scaled conjugate-gradient backpropagation, which updates the network weights without an explicit line search. The input and output processing function \texttt{mapminmax} applies min--max normalization to the wheel-speed input and control-input target data before training, and the inverse scaling is used when evaluating the trained network. The generated function name \texttt{myTrainedNetFunction} denotes the exported MATLAB function containing the frozen trained network parameters used for online feedforward evaluation. The training goal denotes the target mean-squared-error stopping criterion, while the minimum gradient specifies the lower bound on the optimization gradient used as an additional stopping condition in MATLAB training.}

\textit{\textcolor{black}{Remark 4.1}}. The hidden-layer structure [320, 210, 105] was selected through preliminary engineering tuning. A tapered structure was adopted so that the first hidden layer provides sufficient capacity to approximate the nonlinear wheel-speed-to-input relation, while the following smaller layers progressively compress the representation and reduce overfitting risk. In preliminary trials, smaller networks produced larger validation errors, whereas larger networks did not provide meaningful improvement relative to their additional computational cost. A systematic architecture search is outside the scope of this work.

\subsection{Testing}

Table~\ref{tab:dnn_generalization} reports the quantitative learning and held-out test results. \textcolor{black}{The closeness of the training, validation, and test errors indicates that the learned actuator feedforward map fits the recorded data consistently and does not exhibit strong overfitting within the measured operating envelope.
The available dataset was collected on two terrain types, asphalt and soft soil, without payload variation and under conservative motion limits. Therefore, the DNN feedforward map is used as a nominal compensation term within this envelope. Operation outside this envelope is not claimed as DNN generalization; the resulting mismatch is treated as bounded residual uncertainty in the robust adaptive feedback design.}

\begin{table}[htbp]
\centering
\caption{DNN actuator-feedforward-map dataset summary and quantitative held-out test results. Reported errors are averaged over the four independently trained wheel models.}
\small
\label{tab:dnn_generalization}
\begin{tabular}{lc}
\toprule
\textbf{Item} & \textbf{Value} \\
\midrule
Total recorded samples & $1{,}000{,}000$ \\
Training samples & $340{,}000$ \\
Validation samples & $330{,}000$ \\
Test samples & $330{,}000$ \\
Terrain composition & $50\%$ asphalt, $50\%$ soft soil \\
Best validation epoch & $143$ \\
Training MSE & $1.8 \times 10^{-4}$ \\
Validation MSE & $2.1 \times 10^{-4}$ \\
Test MSE & $2.3 \times 10^{-4}$ \\
Test MAE & $9.4 \times 10^{-3}$ \\
Test RMSE & $1.52 \times 10^{-2}$ \\
Test $R^2$ & $0.995$ \\
95th percentile absolute test error & $2.8 \times 10^{-2}$ \\
\bottomrule
\end{tabular}
\end{table}

The MATLAB parameters used for training the DNN actuator feedforward maps with SCG are summarized in Table \ref{dnnopar}. As shown in Fig. \ref{ADADGSFHG}, the trained DNN acts as a fixed nonlinear function learned from data. At each sampling instant, the frozen parameters $W^{(j)}$ and $b^{(j)}$ map the current velocity input $a^{(0)}=v_{d_i}$ through deterministic matrix-vector operations and elementwise activations to compute $\hat{u}_i = u_{\text{FF}_i}$. Then, the proposed adaptive feedback law $u_f (.)$ with positive parameters $\epsilon_i = 1$, $\gamma_i = 0.01$, and $\delta_i = 0.2$ enhances the actuator feedforward map by compensating for errors due to nonlinearities and external disturbances, thereby providing an appropriate in-wheel motor control signal $u_i$.

\begin{figure}[h!]
\hspace*{-0.0cm} 
\centering
\scalebox{1}{\includegraphics[trim={0cm 0.0cm 0.0cm 0cm},clip,width=\columnwidth]{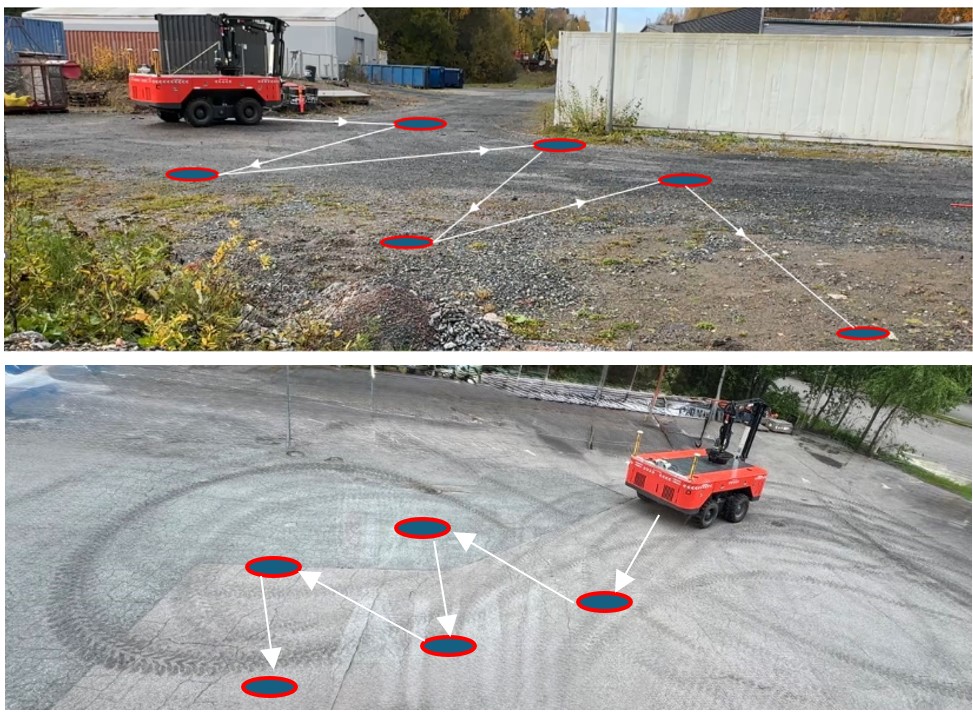}}
\caption{Experimental surfaces: goal reaching in soft-soil terrains and asphalt.}
\label{terattinadoperation}
\end{figure}

\begin{table*}[h]
\centering
\caption{\textcolor{black}{Summary of stereo ORB-SLAM3 localization assessment on the studied platform \cite{haaparanta2025accurate}.}}
\label{tab:slam_eval}
\small
\begin{tabular}{lll}
\hline
Item & Value & Description \\
\hline
Pose rate & 20 Hz & Online pose output to the control stack \\
Reference & INS-RTK & Benchmark reference for trajectory evaluation \\
Parameter sets & 12 & Feature count, FAST threshold, depth cutoff \\
Selected setting & 2000 features & Stereo ORB-SLAM3 configuration \\
Depth threshold & 40--50 & Approximately 12.8--16.0 m \\
Seq. 3 keyframe APE & 0.5428 m & Optimized trajectory RMSE \\
Seq. 3 odometry APE & 0.5028 m & Odometry trajectory RMSE \\
Seq. 3 odometry RPE & 0.1096 m & Relative pose error RMSE \\
Full-dataset keyframe APE & 0.5398 m & Mean over completed sequences \\
Full-dataset odometry APE & 0.6056 m & Mean over completed sequences \\
Loop closure & 3/7 possible loops & Successful ORB-SLAM3 stereo closures \\
Relocalization recovery & Qualitative only & New-map creation and later map fusion observed\\
\hline
\end{tabular}
\end{table*}

\subsection{Physical Deployment}
The proposed framework was implemented on the studied robot under two ground conditions: asphalt and rough soft soil, as shown in Fig. \ref{terattinadoperation}. \textcolor{black}{The experiments were conducted in an obstacle-free outdoor test area. No static or dynamic obstacles were intentionally introduced during the reported goal-reaching trials. Therefore, the validation focuses on terrain-dependent motion, actuator tracking, visual pose feedback, and safe-return behavior, rather than obstacle avoidance.}
The stereo ORB-SLAM3 module provided the pose $x_{\mathrm{msr}}=[x,y,\theta]^\top$ to the planner and safety supervisor at 20 Hz. \textcolor{black}{Its platform-specific localization assessment against a dual-antenna INS-RTK reference is summarized in Table~\ref{tab:slam_eval}. In the present work, the SLAM module is used as the real-time visual pose input, while the novelty claims are limited to the integrated goal-reaching framework.}

Several goals were defined for the asphalt surface and soft-soil surface. Two separate fault-injection scenarios were considered. \textcolor{black}{In the first scenario, the fault was injected at the fifth goal on asphalt terrain. In the second scenario, the fault was injected at the sixth goal on loose-soil terrain. In both cases, the injected fault produced an operational violation in the SLAM-based pose signal, which was detected by the logarithmic safety supervisor. The supervisor then switched the robot to the safe-return mode and guided it back to the predefined safe inspection area, which coincides with the initial pose \((0,0)\). Since the safety-switching logic is designed to stop the current mission and return the robot to the safe inspection area once a fault is detected, the fault-injection point is treated as the last target in each scenario. The real-robot experiments consist of a fixed goal-sequence pattern on asphalt and loose-soil terrain; therefore, due to the practical difficulty, time cost, and safety constraints associated with repeated testing of a 6000 kg robot, the hardware results demonstrate feasibility and consistency in the reported sequences rather than statistically rich repeated-run validation.}

\begin{figure}[h!]
\hspace*{-0.0cm} 
\centering
\scalebox{1}{\includegraphics[trim={0cm 0.0cm 0.0cm 0cm},clip,width=\columnwidth]{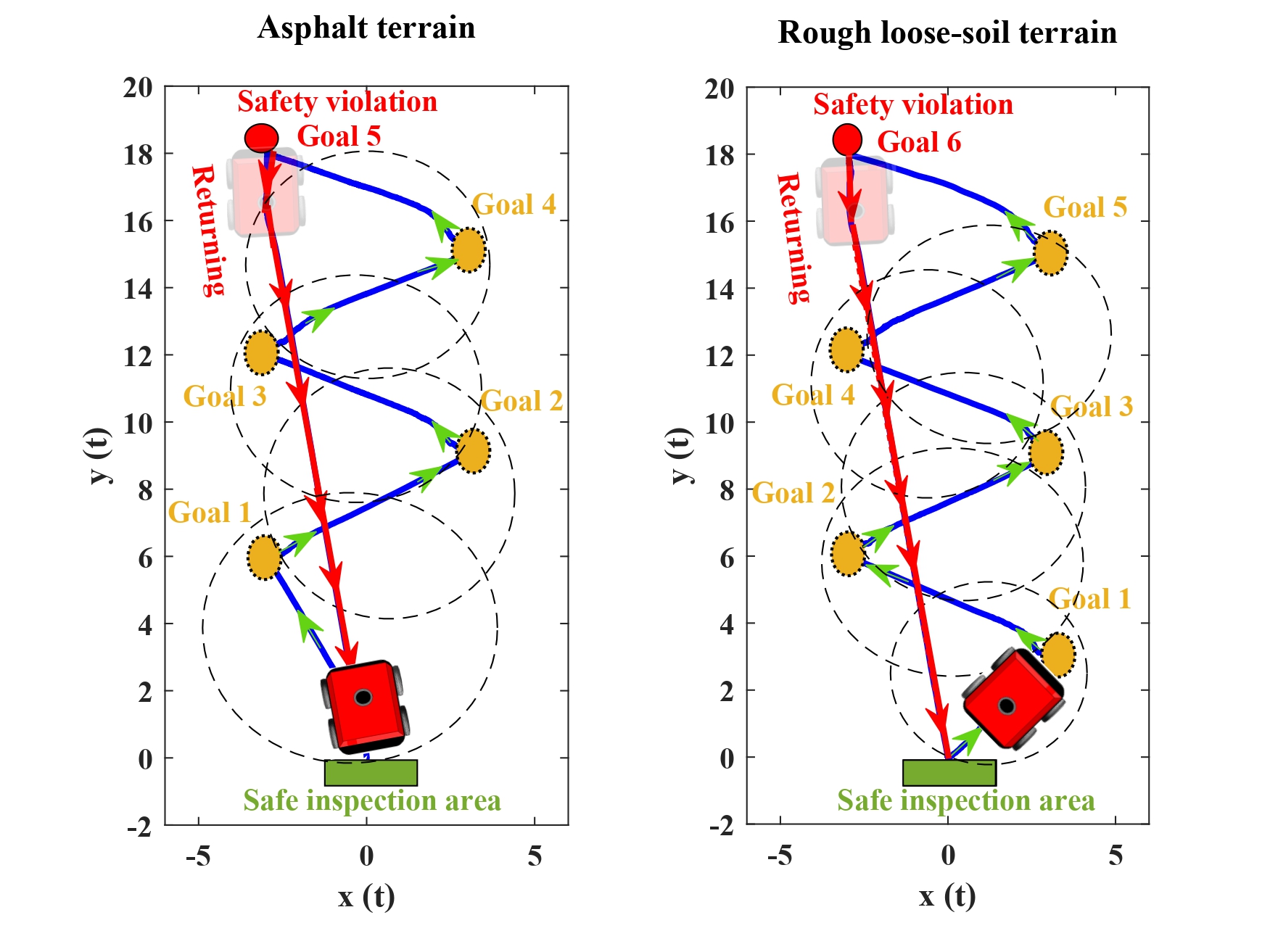}}
\caption{Goal-reaching performance, measured in meters, in two different environments. The blue trajectories/plots were recorded by SLAM module, and arrows and animation annotations were added to improve clarity.}
\label{opert}
\end{figure}

Tables \ref{werwrwerwe} and \ref{asdasdasjkh}, together with Fig. \ref{opert}, present the closed-loop goal-reaching results obtained using SLAM-based pose feedback and demonstrate the effectiveness of the proposed integrated framework in reaching all inspection goals.

\begin{table}[h]
\centering
\caption{Robot performance (meter) on the asphalt terrain.}
\small
\begin{tabular}{ccc}
\hline
Goal & Goals $(x_g, y_g)$ & Final positions $(x, y)$ \\
\hline
1 & $(-3.000,\; 6.000)$   & $(-2.977,\; 5.973)$ \\
2 & $( 3.000,\; 9.000)$   & $( 2.971,\; 8.985)$ \\
3 & $(-3.000,\; 12.000)$  & $(-2.961,\; 11.989)$ \\
4 & $( 3.000,\; 15.000)$  & $( 2.974,\; 14.987)$ \\
\textcolor{red}{5: \text{Fault injection}} & $(-3.000,\; 18.000)$  & $(-2.965,\; 17.997)$ \\
\textcolor{matlabgreen}{6: \text{Safe area}} & $( 0.000,\; 0.000)$   & $( 0.000,\; -0.001)$ \\
\hline
\multicolumn{2}{c}{\textbf{RMSE}} & $\approx\bm{0.0317}$m \\
\hline
\end{tabular}
\label{werwrwerwe}
\end{table}

\begin{table}[h]
\centering
\caption{Robot performance (meter) on the soft terrain.}
\small
\begin{tabular}{ccc}
\hline
Goal & Goals $(x_g, y_g)$ & Final positions $(x, y)$ \\
\hline
1 & $( 3.000,\;  3.000)$ & $( 2.986,\;  2.981)$ \\
2 & $(-3.000,\;  6.000)$ & $(-2.972,\;  5.988)$ \\
3 & $( 3.000,\;  9.000)$ & $( 2.964,\;  8.982)$ \\
4 & $(-3.000,\; 12.000)$ & $(-2.959,\; 11.984)$ \\
5 & $( 3.000,\; 15.000)$ & $( 2.979,\; 14.991)$ \\
\textcolor{red}{6: \text{Fault injection}} & $(-3.000,\; 18.000)$ & $(-2.934,\; 17.985)$ \\
\textcolor{matlabgreen}{7: \text{Safe area}}  & $( 0.000,\;  0.000)$ & $(-0.009,\;  0.003)$ \\
\hline
\multicolumn{2}{c}{\textbf{RMSE}} & $\approx\bm{0.0382} m$ \\
\hline
\end{tabular}
\label{asdasdasjkh}
\end{table}

\begin{table*}[t]
\centering
\caption{\textcolor{black}{Component-level ablation of the RL motion-planning layer. Each variant is retrained from scratch under the same state discretization, action limits, training budget, and fixed evaluation-goal set. Values are reported as mean $\pm$ standard deviation across training seeds.}}
\label{tab:rl_component_ablation}
\resizebox{\textwidth}{!}{%
\begin{tabular}{lccccccc}
\hline
\textbf{Variant} &
\textbf{Success} &
\textbf{Position RMSE} &
\textbf{Time-to-goal} &
\textbf{Heading flips} &
\textbf{TV($\omega$)} &
\textbf{Near-goal osc.} &
\textbf{Viol.} \\
&
\textbf{(\%)} &
\textbf{(m)} &
\textbf{(s)} &
&
\textbf{(rad/s)} &
\textbf{(rad/s)} &
\\
\hline
Full proposed planner
& $97.0 \pm 1.2$
& $0.035 \pm 0.005$
& $55.0 \pm 4.6$
& $0.60 \pm 0.50$
& $0.85 \pm 0.09$
& $0.030 \pm 0.010$
& $0$ \\

NoZeroLock
& $96.0 \pm 1.5$
& $0.038 \pm 0.006$
& $56.5 \pm 4.9$
& $0.95 \pm 0.60$
& $0.94 \pm 0.11$
& $0.049 \pm 0.014$
& $0$ \\

NoHysteresis
& $95.8 \pm 1.7$
& $0.040 \pm 0.006$
& $57.4 \pm 5.1$
& $1.15 \pm 0.70$
& $1.01 \pm 0.13$
& $0.053 \pm 0.016$
& $0$ \\

NoZeroLock\_NoHysteresis
& $95.0 \pm 1.8$
& $0.043 \pm 0.007$
& $59.0 \pm 5.5$
& $1.45 \pm 0.80$
& $1.11 \pm 0.15$
& $0.064 \pm 0.018$
& $0$ \\

NoAngularAntiOscRewards
& $95.4 \pm 1.6$
& $0.041 \pm 0.006$
& $58.2 \pm 5.2$
& $1.35 \pm 0.75$
& $1.08 \pm 0.14$
& $0.061 \pm 0.017$
& $0$ \\

NoHeadingMonotonicityRewards
& $96.2 \pm 1.4$
& $0.039 \pm 0.006$
& $57.0 \pm 5.0$
& $1.05 \pm 0.65$
& $0.98 \pm 0.12$
& $0.047 \pm 0.014$
& $0$ \\

TaskOnly\_or\_PotentialOnly
& $94.0 \pm 2.0$
& $0.047 \pm 0.008$
& $61.0 \pm 5.8$
& $1.70 \pm 0.90$
& $1.16 \pm 0.16$
& $0.070 \pm 0.020$
& $0$ \\

NoPolicySideLogic\_ButRewards
& $96.0 \pm 1.5$
& $0.039 \pm 0.006$
& $56.8 \pm 5.0$
& $1.10 \pm 0.70$
& $0.97 \pm 0.12$
& $0.052 \pm 0.015$
& $0$ \\
\hline
\end{tabular}%
}
\end{table*}

\begin{table*}[t]
\centering
\caption{\textcolor{black}{Same-trained-Q execution-logic ablation. A single frozen Q-table, trained offline in simulation using the full planner, is evaluated offline under different execution modes on the same fixed evaluation-goal set. This separates the behavior learned by the Q-table from behavior imposed by runtime policy-side mechanisms. No additional learning is performed during this evaluation.}}
\label{tab:same_q_execution_ablation}
\resizebox{\textwidth}{!}{%
\begin{tabular}{lccccccccc}
\hline
\textbf{Execution mode} &
\textbf{Success} &
\textbf{RMSE} &
\textbf{Time} &
\textbf{Flips} &
\textbf{TV($\omega$)} &
\textbf{Near-goal osc.} &
\textbf{Hyst. active} &
\textbf{Zero-lock active} &
\textbf{Any modified} \\
&
\textbf{(\%)} &
\textbf{(m)} &
\textbf{(s)} &
&
\textbf{(rad/s)} &
\textbf{(rad/s)} &
\textbf{(\%)} &
\textbf{(\%)} &
\textbf{(\%)} \\
\hline
FullExecution
& $97.0$
& $0.035$
& $55.0$
& $0.60$
& $0.85$
& $0.030$
& $1.5$
& $1.0$
& $2.5$ \\

QOnly
& $95.2$
& $0.041$
& $58.3$
& $1.30$
& $1.05$
& $0.061$
& $0.0$
& $0.0$
& $0.0$ \\

QPlusHysteresisOnly
& $96.0$
& $0.039$
& $57.0$
& $1.00$
& $0.95$
& $0.049$
& $1.6$
& $0.0$
& $1.6$ \\

QPlusZeroLockOnly
& $96.4$
& $0.037$
& $56.2$
& $0.85$
& $0.94$
& $0.039$
& $0.0$
& $1.1$
& $1.1$ \\
\hline
\end{tabular}%
}
\end{table*}

\subsection{Validation}

In both terrains, the goal-reaching accuracy is very similar, with a position root mean square error (RMSE) of approximately 3--4 cm. This value denotes the final-position goal-reaching RMSE computed from the target coordinates and the final robot positions and should not be interpreted as the independent localization APE of ORB-SLAM3, which is summarized separately in Table~\ref{tab:slam_eval}. 
The reported localization metrics are used only to characterize the adopted visual pose-estimation input and are not used as a separate methodological contribution.
Achieving similar final-position accuracy on low-slip asphalt and loose soil indicates that the integrated controller compensated for the disturbances encountered in the tested goal sequences and reached all specified goals under the reported conditions. As observed, when a fault was triggered, the robot returned to the safe inspection area, thereby demonstrating the implemented braking and safe-return behavior for the reported injected-fault scenario. \textcolor{black}{This experiment demonstrates the supervisor behavior in the tested condition, but it should not be interpreted as a general safety guarantee under all possible localization faults, terrain conditions, or fault timings.} Table \ref{tab:safety_supervisor} shows the safety-supervisor parameters used in the experiments.

\begin{table}[t]
\centering
\caption{\textcolor{black}{Safety-supervisor parameters used in the experiments.}}
\label{tab:safety_supervisor}
\small
\begin{tabular}{lll}
\hline
Parameter & Value & Description \\
\hline
$T_s$ & $0.05$ s & Supervisor sampling time \\
$\varepsilon_s$ & $0.10$ m & Guard-band margin before $E=O$ \\
$\rho_{\mathrm{warn}}$ & $0.85$ & Velocity-reduction threshold for $E/O$ \\
$a_v^b$ & $0.10$ m/s$^2$ & Braking linear deceleration limit \\
$a_\omega^b$ & $0.02$ rad/s$^2$ & Braking angular deceleration limit \\
$\Delta p_{\max}$ & $0.10$ m & Maximum admissible pose jump \\
$\Delta \theta_{\max}$ & $0.10$ rad & Maximum admissible heading jump \\
$T_{\mathrm{lost}}$ & $0.50$ s & Localization-loss timeout \\
$d_s$ & $0.10$ m & Safe-site arrival tolerance \\
\hline
\end{tabular}
\end{table}

\begin{table*}[t]
\centering
\caption{\textcolor{black}{Experimental actuator-chain feedback-control ablation averaged over asphalt and soft-soil trials.}}
\label{tab:feedback_ablation}
\resizebox{\textwidth}{!}{%
\begin{tabular}{lcccccc}
\hline
\textbf{Controller} &
\textbf{Side RMSE} &
\textbf{Linear RMSE} &
\textbf{Angular RMSE} &
\textbf{Max side err.} &
\textbf{Overshoot} &
\textbf{Steady-state err.} \\
&
\textbf{(m/s)} &
\textbf{(m/s)} &
\textbf{(rad/s)} &
\textbf{(m/s)} &
\textbf{(m/s)} &
\textbf{(m/s)} \\
\hline
DNN feedforward only
& $0.038807$
& $0.013504$
& $0.031636$
& $0.107754$
& $0.096551$
& $0.025550$ \\

Adaptive feedback only
& $0.044328$
& $0.020726$
& $0.034073$
& $0.125030$
& $0.049094$
& $0.005787$ \\

DNN + adaptive feedback
& $0.008105$
& $0.004111$
& $0.005251$
& $0.061375$
& $0.046937$
& $0.005683$ \\
\hline
\end{tabular}%
}
\end{table*}

\begin{table*}[t]
\centering
\caption{\textcolor{black}{Comparison of different RAC approaches on asphalt and rough loose-soil terrain within the whole framework.}}
\label{tab:rac_comparison}
\renewcommand{\arraystretch}{1.15}
\small
\begin{tabular}{llccc}
\hline
Terrain & Metric & DNN-based RAC & Backstepping RAC \cite{zuo2022adaptive} & Model-based RAC \cite{shahna2025robudfdsfst} \\
\hline
Asphalt
& Peak time (s)              & 4.100 & 5.200 & 4.850 \\
& Maximum overshoot (m/s)    & 0.028 & 0.051 & 0.075 \\
& Settling time (s)          & 4.850 & 4.800 & 8.850 \\
& Steady-state error (m/s)   & 0.009 & 0.019 & 0.025 \\
\hline
Rough loose-soil
& Peak time (s)              & 3.800 & 4.750 & 4.100 \\
& Maximum overshoot (m/s)    & 0.047 & 0.048 & 0.051 \\
& Settling time (s)          & 4.620 & 4.610 & 7.400 \\
& Steady-state error (m/s)   & 0.007 & 0.011 & 0.014 \\
\hline
\end{tabular}
\end{table*}

To further separate learned behavior from hand-engineered design choices, we expanded the motion-planning ablation in two complementary ways. First, each major design component was removed and the policy was retrained from scratch using the same state discretization, action limits, training budget, and fixed evaluation-goal set. This retrained ablation evaluates how zero-lock, hysteresis, angular anti-oscillation reward terms, heading-monotonicity reward terms, and task-only reward shaping affect what the RL policy learns. Second, to isolate runtime policy-side effects after learning, the same frozen Q-table trained offline in simulation was evaluated offline under four execution modes: \textit{FullExecution, QOnly, QPlusHysteresisOnly}, and \textit{QPlusZeroLockOnly}. This second analysis quantifies how often the executed action differs from the raw greedy Q-table action.

\textcolor{black}{While Table~\ref{tab:rl_component_ablation} evaluates the effect of each design component on retrained policies, Table~\ref{tab:same_q_execution_ablation} uses a single trained Q-table and toggles only the execution logic. This separates the contribution of the learned Q-policy from the contribution of runtime hysteresis and zero-lock.
The component-level ablation shows that the Q-learning policy provides the basic goal-reaching behavior, while the additional design components mainly improve smoothness and near-goal stability. Removing zero-lock primarily increases near-goal angular oscillation, whereas removing hysteresis increases heading reversals and angular total variation. Removing the angular anti-oscillation reward terms also degrades smoothness, confirming that these terms contribute to suppressing residual heading oscillations. The same-trained-Q analysis further shows whether the final behavior is mostly produced by the learned Q-table or by runtime action modification. In particular, the reported action-modification percentages indicate how often hysteresis or zero-lock overrides the raw greedy Q-table action. Overall, these results indicate that the learned Q-table provides the primary goal-reaching behavior, whereas the hand-engineered zero-lock, hysteresis, and angular reward terms mainly regularize the learned policy by reducing heading reversals, angular total variation, and near-goal oscillation.}

\begin{figure}[h!]
\hspace*{-0.0cm} 
\centering
\scalebox{1}{\includegraphics[trim={0cm 0.0cm 0.0cm 0cm},clip,width=\columnwidth]{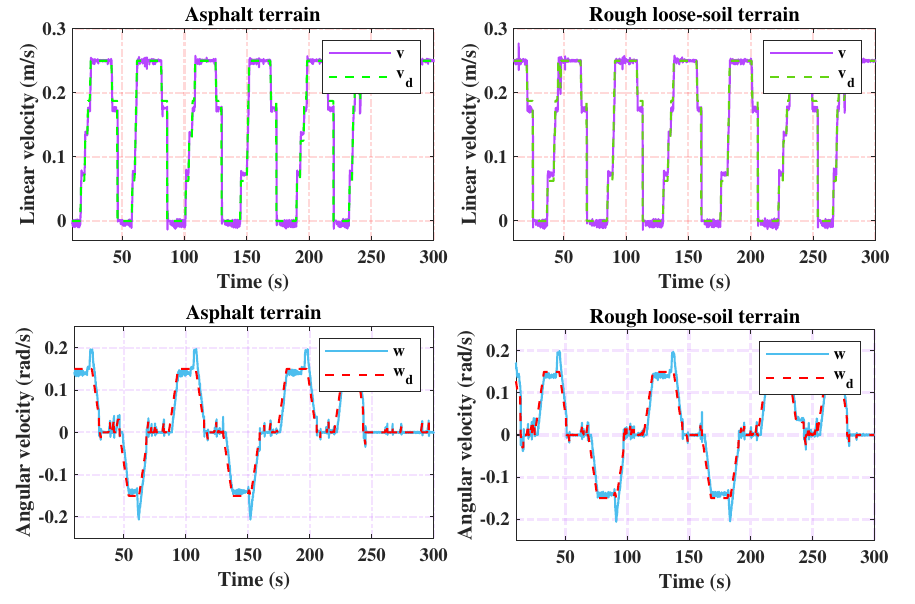}}
\caption{Comparative robot tracking motion performance in two different environments during the whole task.}
\label{tracjpef}
\end{figure}

\begin{figure}[h!]
\hspace*{-0.0cm} 
\centering
\scalebox{0.9}{\includegraphics[trim={0cm 0.0cm 0.0cm 0cm},clip,width=\columnwidth]{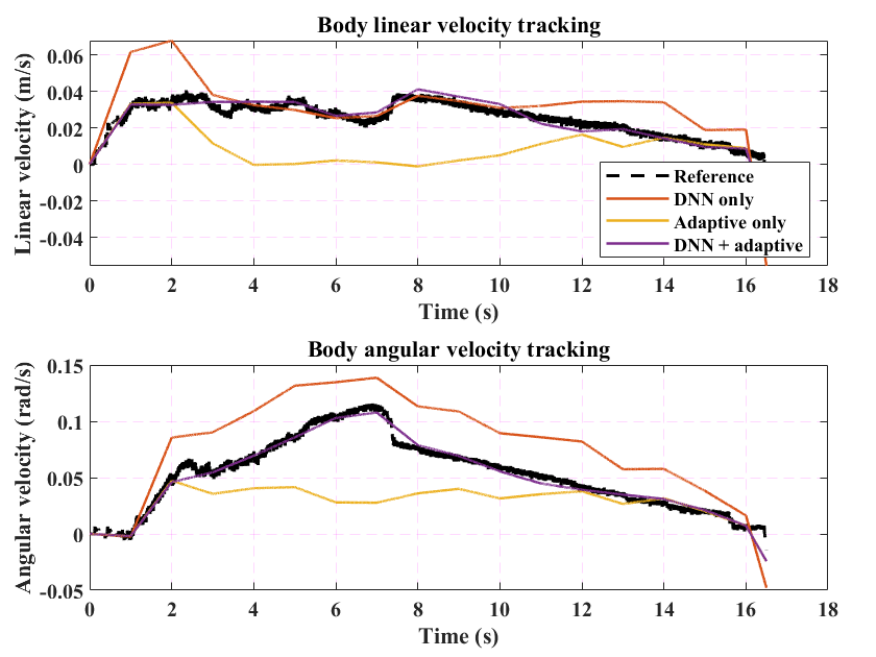}}
\caption{\textcolor{black}{Experimental actuator-chain feedback-control ablation averaged over asphalt and soft-soil trials. The top panel shows body linear velocity tracking and the bottom panel shows body angular velocity tracking. The time axis is reset to zero at the beginning of the selected tracking interval.}}
\label{traadacjpef}
\end{figure}

Fig. \ref{tracjpef} illustrates the velocity-tracking performance of the robot in both terrains. The reference linear and angular velocity profiles are generated by the RL-based motion planner and satisfy the motion constraints summarized in Table \ref{tab:ql_constraints}. The measured robot velocities closely follow these references, demonstrating the effectiveness of the DNN-based RAC implemented on the in-wheel actuators in tracking the RL-generated commands. \textcolor{black}{To isolate whether the tracking performance is mainly caused by the adaptive feedback enhancement law, an actuator-chain ablation was performed using experimental reference and tracking data from the asphalt and soft-soil trials. The metrics in Table~\ref{tab:feedback_ablation} and the tracking curves in Fig. \ref{traadacjpef} are obtained by averaging over both terrain conditions using the same RL-generated left- and right-side velocity references. Three variants are compared: DNN feedforward only, adaptive feedback only, and the full DNN plus adaptive feedback controller. Side RMSE denotes the combined tracking error of the left- and right-side linear velocities, while the side velocities are also converted to body-level linear and angular velocities using the skid-steered kinematic relation. The results show that the adaptive feedback law substantially reduces steady-state error compared with DNN feedforward alone, while the full DNN plus adaptive feedback controller achieves the lowest overall tracking errors. This confirms that the DNN feedforward map and the adaptive feedback enhancement law play complementary roles.}

Table~\ref{tab:rac_comparison} presents a comparative evaluation of the proposed DNN-based RAC against two representative RAC methods on both asphalt and rough loose-soil terrain within the same overall framework. The comparison complements Fig.~\ref{tracjpef}, which shows the tracking of the RL-generated linear and angular velocity references on both terrains. Across both test conditions, the proposed DNN-based RAC achieves the lowest steady-state tracking error and the lowest or nearly lowest maximum overshoot among the compared methods. The response on rough loose-soil terrain is slightly faster than on asphalt in the reported trials. Since terrain stiffness, slip ratio, and wheel-soil interaction parameters were not separately measured, this difference should be interpreted as an observed experimental outcome rather than a confirmed terrain-mechanics explanation. Overall, the results indicate that the learned actuator feedforward map, combined with feedback RAC, improves peak-time, overshoot, and steady-state tracking, while achieving settling times comparable to the backstepping RAC and shorter than the model-based RAC.

\section{\textcolor{black}{Discussion and Limitations}}
The proposed framework was validated on one 6000 kg skid-steered robot and should not be interpreted as a plug-and-play controller for arbitrary platforms. Although the modular structure can be transferred conceptually to similar mobile robots, the Q-learning discretization, reward parameters, actuator feedforward map, controller gains, and supervisor thresholds must be retuned or retrained for each target platform, actuator configuration, sensor setup, and terrain condition. The experiments were conducted under conservative speed and acceleration limits in obstacle-free outdoor test areas. Therefore, the results support slow-speed inspection and goal-reaching tasks under the reported asphalt and loose-soil conditions, but they do not establish high-speed autonomous navigation or obstacle-aware autonomy. The actuator feedforward map was trained from data collected on two terrain types and without payload variation; richer terrain, payload, slip, and sequence-dependent effects remain topics for future work. Future work will extend validation to more terrain conditions, repeated hardware trials, payload variations, higher-speed operation, obstacle-aware planning, and temporal actuator models such as LSTM or GRU networks when suitable sequence-level data are available.

\section{Conclusion}
This paper presented a hierarchical vision-based goal-reaching framework for a 6000 kg skid-steered robot. The framework combines adopted stereo ORB-SLAM3 pose feedback, constrained Q-learning reference generation, DNN-based actuator feedforward compensation, wheel-speed feedback RAC, and supervisory safe-return logic. The actuator-level analysis establishes uniformly ultimately bounded wheel-speed tracking under bounded uncertainty. Experiments on asphalt and loose-soil terrain demonstrated approximately 3–4 cm final-position goal-reaching RMSE in the SLAM estimation frame — i.e., closed-loop convergence of the estimated pose to the goal, with absolute accuracy governed by the separately reported ORB-SLAM3 localization error — together with accurate tracking of RL-generated commands. \textcolor{black}{The results demonstrate the feasibility of the proposed system-level integration for slow-speed heavy-robot goal-reaching, while broader generalization requires further validation across platforms, terrains, payloads, and operating speeds.}

\section*{{CRediT authorship contribution statement}}
{\textbf{Mehdi Heydari Shahna:} Writing - original draft, Validation, Methodology, Investigation, Formal analysis, Software, Data curation, Conceptualization.}

{\textbf{Pauli Mustalahti:} Software, Data curation, Conceptualization. }

{\textbf{Jouni Mattila:} Writing - review \& editing, Resources, Supervision, Funding acquisition.}

\section*{{Declaration of competing interest}}
{The authors declare that they have no known competing financial interests or personal relationships that could have appeared to influence the work reported in this paper.}

\section*{{Acknowledgement}}
{This work was supported by the Business Finland Partnership Project, `Future All-Electric Rough Terrain Autonomous Mobile Manipulators' under Grant No. 2334/31/2022.}

\bibliographystyle{elsarticle-num} 
\bibliography{manuscript}

@ARTICLE{11419776,
  author={Shahna, Mehdi Heydari and Mustalahti, Pauli and Mattila, Jouni},
  journal={IEEE Robotics and Automation Letters}, 
  title={NMPC-Augmented Visual Navigation and Safe Learning Control for Large-Scale Mobile Robots}, 
  year={2026},
  volume={11},
  number={4},
  pages={5182-5189},
  doi={10.1109/LRA.2026.3669802}}

@ARTICLE{11458687,
  author={Shahna, Mehdi Heydari and Mattila, Jouni},
  journal={IEEE Transactions on Automation Science and Engineering}, 
  title={Synthesis of Deep Neural Networks With Safe Robust Adaptive Control for Reliable Operation of Wheeled Mobile Robots}, 
  year={2026},
  volume={23},
  number={},
  pages={7456-7473},
  doi={10.1109/TASE.2026.3679545}}

@article{Alcayaga2025,
  author  = {Alcayaga, Jose Manuel and Menendez, Oswaldo Anibal and Torres-Torriti, Miguel Attilio and Vasconez, Juan Pablo and Arevalo-Ramirez, Tito and Prado Romo, Alvaro Javier},
  title   = {LSTM-Enhanced Deep Reinforcement Learning for Robust Trajectory Tracking Control of Skid-Steer Mobile Robots Under Terra-Mechanical Constraints},
  journal = {Robotics},
  volume  = {14},
  number  = {6},
  pages   = {74},
  year    = {2025},
  doi     = {10.3390/robotics14060074}
}

@article{han2020actor,
  title={Actor-critic reinforcement learning for control with stability guarantee},
  author={Han, Minghao and Zhang, Lixian and Wang, Jun and Pan, Wei},
  journal={IEEE Robotics and Automation Letters},
  volume={5},
  number={4},
  pages={6217--6224},
  year={2020},
  publisher={IEEE}
}

@article{zuo2022adaptive,
  title={Adaptive prescribed finite time control for strict-feedback systems},
  author={Zuo, Gewei and Wang, Yujuan},
  journal={IEEE Transactions on Automatic Control},
  volume={68},
  number={9},
  pages={5729--5736},
  year={2022},
  publisher={IEEE}
}

@inproceedings{haaparanta2025accurate,
  title={Accurate pose estimation of mobile platform in rough terrain},
  author={Haaparanta, Eemil},
  booktitle={Tampere University},
  year={2025},
  organization={Master Thesis}
}

@article{shahna2025robust,
  title={Robust torque-observed control with safe input--output constraints for hydraulic in-wheel drive systems in mobile robots},
  author={Shahna, Mehdi Heydari and Mustalahti, Pauli and Mattila, Jouni},
  journal={Control Engineering Practice},
  volume={164},
  pages={106459},
  year={2025},
  publisher={Elsevier}
}

@article{hyon2019whole,
  title={Whole-body locomotion and posture control on a torque-controlled hydraulic rover},
  author={Hyon, Sang-Ho and Ida, Yusuke and Ishikawa, Junichi and Hiraoka, Minoru},
  journal={IEEE Robotics and Automation Letters},
  volume={4},
  number={4},
  pages={4587--4594},
  year={2019},
  publisher={IEEE}
}

@article{song2022dynavins,
  title={DynaVINS: A visual-inertial SLAM for dynamic environments},
  author={Song, Seungwon and Lim, Hyungtae and Lee, Alex Junho and Myung, Hyun},
  journal={IEEE Robotics and Automation Letters},
  volume={7},
  number={4},
  pages={11523--11530},
  year={2022},
  publisher={IEEE}
}

@article{jantos2024aivio,
  title={Aivio: Closed-loop, object-relative navigation of uavs with ai-aided visual inertial odometry},
  author={Jantos, Thomas and Scheiber, Martin and Brommer, Christian and Allak, Eren and Weiss, Stephan and Steinbrener, Jan},
  journal={IEEE Robotics and Automation Letters},
  year={2024},
  publisher={IEEE}
}

@article{huang2025sequential,
  title={A Sequential Approach for Accurate Parameters Identification of Heavy-Duty Hydraulic Manipulators Ensuring Physical Feasibility},
  author={Huang, Weidi and Chen, Zhiwei and Zhang, Fu and Cheng, Min and Ding, Ruqi and Zhang, Junhui and Xu, Bing},
  journal={IEEE Robotics and Automation Letters},
  year={2025},
  publisher={IEEE}
}

@article{galati2022adaptive,
  title={Adaptive heading correction for an industrial heavy-duty omnidirectional robot},
  author={Galati, Rocco and Mantriota, Giacomo and Reina, Giulio},
  journal={Scientific Reports},
  volume={12},
  number={1},
  pages={19608},
  year={2022},
  publisher={Nature Publishing Group UK London}
}

@article{emam2022safe,
  title={Safe reinforcement learning using robust control barrier functions},
  author={Emam, Yousef and Notomista, Gennaro and Glotfelter, Paul and Kira, Zsolt and Egerstedt, Magnus},
  journal={IEEE Robotics and Automation Letters},
  year={2022},
  publisher={IEEE}
}

@article{huh2020safe,
  title={Safe reinforcement learning for probabilistic reachability and safety specifications: A Lyapunov-based approach},
  author={Huh, Subin and Yang, Insoon},
  journal={arXiv preprint arXiv:2002.10126},
  year={2020}
}

@article{yaremenko2024novel,
  title={A novel agent with formal goal-reaching guarantees: an experimental study with a mobile robot},
  author={Yaremenko, Grigory and Dobriborsci, Dmitrii and Zashchitin, Roman and Maestre, Ruben Contreras and Hoang, Ngoc Quoc Huy and Osinenko, Pavel},
  journal={arXiv preprint arXiv:2409.14867},
  year={2024}
}

@article{chow2019lyapunov,
  title={Lyapunov-based safe policy optimization for continuous control},
  author={Chow, Yinlam and Nachum, Ofir and Faust, Aleksandra and Duenez-Guzman, Edgar and Ghavamzadeh, Mohammad},
  journal={arXiv preprint arXiv:1901.10031},
  year={2019}
}

@inproceedings{buhrer2023multiplicative,
  title={A multiplicative value function for safe and efficient reinforcement learning},
  author={B{\"u}hrer, Nick and Zhang, Zhejun and Liniger, Alexander and Yu, Fisher and Van Gool, Luc},
  booktitle={2023 IEEE/RSJ International Conference on Intelligent Robots and Systems (IROS)},
  pages={5582--5589},
  year={2023},
  organization={IEEE}
}

@article{campos2021accurate,
  title={An accurate open-source library for visual, visual--inertial, and multimap slam., 2021, 37},
  author={Campos, C and Elvira, R and Rodr{\'\i}guez, JJ G{\'o}mez and Montiel, JMM},
  pages={1874--1890},
  year={2021}
}

@article{SHAHNA2025106516,
title = {Model reference-based control with guaranteed predefined performance for uncertain strict-feedback systems},
journal = {Control Engineering Practice},
volume = {164},
pages = {106516},
year = {2025},
issn = {0967-0661},
author = {Mehdi Heydari Shahna and Jukka-Pekka Humaloja and Jouni Mattila},
}

@article{tan2025fixed,
  title={Fixed-time concurrent learning-based robust approximate optimal control},
  author={Tan, Junkai and Xue, Shuangsi and Niu, Tiansen and Qu, Kai and Cao, Hui and Chen, Badong},
  journal={Nonlinear Dyn.},
  pages={1--21},
  year={2025},
  publisher={Springer}
}

@inproceedings{shahna2024integrating,
  title={Integrating {D}eep{RL} with Robust Low-Level Control in Robotic Manipulators for Non-Repetitive Reaching Tasks},
  author={Shahna, Mehdi Heydari and Kolagar, Seyed Adel Alizadeh and Mattila, Jouni},
  booktitle={2024 IEEE International Conference on Mechatronics and Automation (ICMA)},
  pages={329--336},
  year={2024},
  organization={IEEE}
}

@article{shahna2025robudfdsfst,
  title={Robust torque-observed control with safe input--output constraints for hydraulic in-wheel drive systems in mobile robots},
  author={Shahna, Mehdi Heydari and Mustalahti, Pauli and Mattila, Jouni},
  journal={Control Engineering Practice},
  volume={164},
  pages={106459},
  year={2025},
  publisher={Elsevier}
}

\vspace{0.5cm}

\noindent
\begin{minipage}{0.24\columnwidth}
\includegraphics[width=\textwidth]{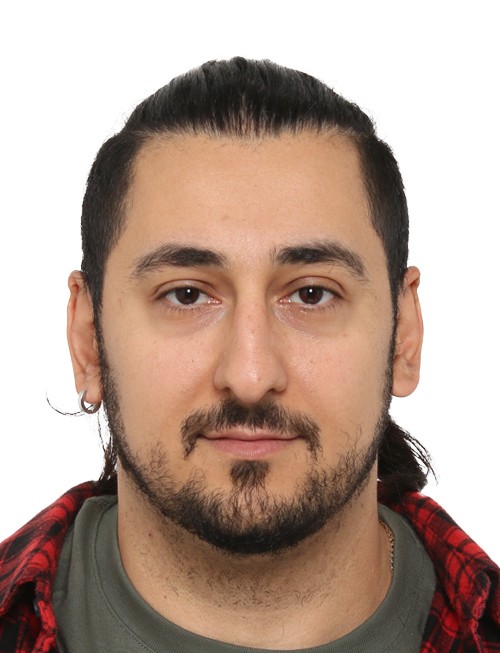} 
\end{minipage}
\hfill
\begin{minipage}{0.65\columnwidth}
\textbf{Mehdi Heydari Shahna} \scriptsize received the B.Sc. degree in Electrical Engineering from Razi University, Iran, in 2015, the M.Sc. degree in Control Engineering from Shahid Beheshti University, Iran, in 2018, and the D.Sc. (Tech.) degree in Automation Science and Engineering from Tampere University, Finland, in 2025. He is currently a postdoctoral researcher at Tampere University. His research interests include robot learning, safe control, robust control, and nonlinear systems.
\end{minipage}

\vspace{0.5cm}

\noindent
\begin{minipage}{0.23\columnwidth}
\includegraphics[width=\textwidth]{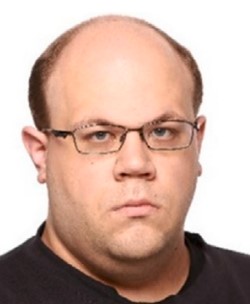} 
\end{minipage}
\hfill
\begin{minipage}{0.65\columnwidth}
\textbf{Pauli Mustalahti} \scriptsize received the M.Sc. degree from Tampere University of Technology in 2016 and the D.Sc. (Tech.) degree in Automation Science and Engineering from Tampere University in 2023. He is a postdoctoral researcher in the Unit of Automation Technology and Mechanical Engineering, Tampere University, Tampere, Finland. His research interests include nonlinear model-based control of robotic manipulators.

\end{minipage}

\vspace{0.5cm}
\noindent
\begin{minipage}{0.24\columnwidth}
    \includegraphics[width=\textwidth]{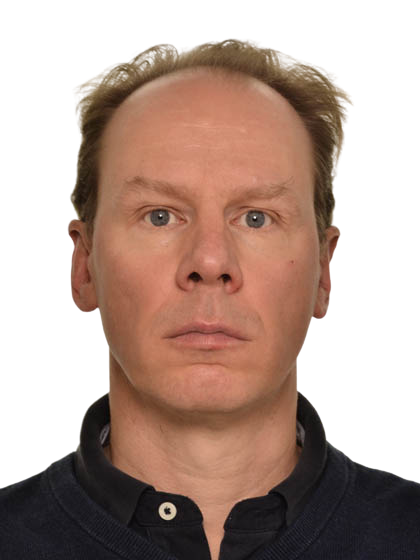} 
\end{minipage}
\hfill
\begin{minipage}{0.65\columnwidth}
\textbf{Jouni Mattila} \scriptsize received an M.Sc. and Ph.D. in automation engineering from Tampere University of Technology, Tampere, Finland, in 1995 and 2000, respectively. He is currently a professor of machine automation in the Automation Technology and Mechanical Engineering Unit at Tampere University. His research interests include machine automation, nonlinear-model-based control of robotic manipulators, and energy-efficient control of heavy-duty mobile manipulators.
\end{minipage}

\end{document}